\documentclass{article} % For LaTeX2e
\usepackage{iclr2023_workshop,times}

\usepackage{amsmath,amsfonts,bm,array,caption}

\newcolumntype{M}[1]{>{\centering\arraybackslash}m{#1}}
\newcolumntype{N}{@{}m{0pt}@{}}

\def\eqref#1{equation~\ref{#1}}
\def\1{\bm{1}}

\def\eps{{\epsilon}}

\DeclareMathAlphabet{\mathsfit}{\encodingdefault}{\sfdefault}{m}{sl}
\SetMathAlphabet{\mathsfit}{bold}{\encodingdefault}{\sfdefault}{bx}{n}

\usepackage{algorithmic}
\usepackage[caption=false]{subfig}
\usepackage[utf8]{inputenc}
\usepackage{blindtext}
\usepackage[a4paper, total={6in, 8in}]{geometry}
\usepackage{amssymb}
\usepackage{amsmath}
\usepackage{amsfonts}
\usepackage{upgreek}
\usepackage{listings}
\usepackage{multirow}
\usepackage{multicol}
\usepackage{bigdelim}
\usepackage{color}
\usepackage{xcolor}
\usepackage{stmaryrd}
\usepackage{soul}
\usepackage[binary-units]{siunitx}
\usepackage[symbol]{footmisc}
\usepackage{booktabs}
\usepackage[section]{placeins} % For '\input{file}'
\usepackage[titletoc]{appendix}
\usepackage{calc}
\usepackage{hyperref}
\usepackage[capitalize]{cleveref}
\usepackage{graphicx}
\usepackage{graphics}
\usepackage{float}
\usepackage{wrapfig}
\usepackage[percent]{overpic}
\usepackage{varwidth}
\usepackage{epstopdf}
\usepackage{adjustbox}
\usepackage{tikz}
\usepackage{pgfplots}
\usetikzlibrary{arrows,matrix,positioning,fit}
\usetikzlibrary{shapes,positioning}
\usetikzlibrary{backgrounds}
\pgfplotsset{compat=1.14}
\usepgfplotslibrary{colorbrewer}
\usepgfplotslibrary{patchplots}
\usepgfplotslibrary[colorbrewer]
\usetikzlibrary{pgfplots.colorbrewer}
\usetikzlibrary[pgfplots.colorbrewer]
\usepgfplotslibrary{fillbetween}
\usepackage{tikz-layers}
\usepgfplotslibrary{units}
\usetikzlibrary{spy}
\usepackage{pgfplotstable}
\graphicspath{ {figs/} }
\usetikzlibrary{external}
\usepackage{enumerate}
\usepackage{enumitem}
\usepackage{comment}

\newcommand{\omegadot}{\boldsymbol{\dot{\omega}}}
\newcommand{\omegaddot}{\boldsymbol{\ddot{\omega}}}
\newcommand{\omegavec}{\boldsymbol{\omega}}
\newcommand{\That}{\mathbf{\hat{T}}}
\newcommand{\V}{\mathbf{V}}
\newcommand{\omegadotnorm}{\|\boldsymbol{\dot\omega}\|_2}
\newcommand{\be}{\begin{equation}}
	\newcommand{\ee}{\end{equation}}
\newcommand{\bea}{\begin{eqnarray}}
	\newcommand{\eea}{\end{eqnarray}}

\newcommand\px[2]{\frac{\partial #1}{\partial {#2}}}

\newcommand\dx[2]{\frac{\mathrm{d} #1}{\mathrm{d} #2}}

\newlength\myheight
\newlength\mydepth
\settototalheight\myheight{Xygp}
\title{Probing optimisation in physics-informed neural networks}

\author{Nayara Fonseca $^*$, Will Trojak $^*$ \\
IBM Research Europe\\
Daresbury, WA44AD, United Kingdom\\
\texttt{\{nayara.fonseca,w.trojak\}@ibm.com} \\ 
\And
Veronica Guidetti $^*$\\
University of Modena and Reggio Emilia \\
Department of Physics, Informatics and Mathematics \\
Via G. Campi 213/a, 41125, Modena, Italy\\
\texttt{veronica.guidetti@unimore.it}
}

\iclrfinalcopy % Uncomment for camera-ready version, but NOT for submission.
\begin{document}

\maketitle

\def\thefootnote{*}\footnotetext{Equal contribution.}\def\thefootnote{\arabic{footnote}}

\begin{abstract}
A novel comparison is presented of the effect of optimiser choice on the accuracy of physics-informed neural networks (PINNs). To give insight into why some optimisers are better, a new approach is proposed that tracks the training trajectory curvature and can be evaluated on the fly at a low computational cost. The linear advection equation is studied for several advective velocities, and we show that the optimiser choice substantially impacts PINNs model performance and accuracy.  Furthermore, using the curvature measure, we found a negative correlation between the convergence error and the curvature in the optimiser local reference frame. It is concluded that, in this case, larger local curvature values result in better solutions. Consequently, optimisation of PINNs is made more difficult as minima are in highly curved regions.

\end{abstract}

% The workshop requires electronic submissions, processed by
% \url{https://openreview.net/}. See the workshop website for more instructions.

% If your paper is ultimately accepted, the statement {\tt
%   {\textbackslash}iclrfinalcopy} should be inserted to adjust the
% format to the camera ready requirements.

\section{Introduction}\label{sec:intro}
\vspace{-0.1cm}  
%\hnf{An attempt to reduce the literature review. All the original text is below btw begin(comments) ... end(comments)}

The idea of solving PDE problems using neural networks (NNs) was put forward by~\cite{Lagaris:1997ap, Lagaris_1998, 870037} in the second half of the '90s and then revised in 2017 by \cite{raissi2017physics_1,raissi2017physics_2} who named the methodology Physics-Informed Neural Networks (PINNs).  Relying on the universal approximation theorem of~\cite{cybenko1989approximation, HORNIK1991251, pinkus_1999}, PINNs aim to deliver a universal regressor that can represent any bounded continuous function and solve any PDE/ODE problems, having input and output shape as the only limitation. Although the goal of PINNs was to produce a  unifying method of solving PDE/ODEs, satisfactory results could not be achieved in a multitude of cases. The recent works of \cite{karniadakis2021physics,hao2022physics} provide an overview of the state-of-the-art; furthermore, \cite{cuomo2022scientific} focus on algorithms and applications and \cite{beck2020overview} on theoretical results.
\vspace{-0.1cm}

Several studies have analysed the effects of choosing different architectures, loss function formulations, and treatment of domain and collocation points. However, the effect of the optimiser choice on PINN performance needs more attention. Recently, new PINN-specific optimisation methods were developed or applied to improve poor convergence performance. For example, \cite{de2022born} propose a \emph{relativistic} optimisation algorithm that introduces chaotic jumps and \cite{davi2022pso} use a metaheuristic optimisation method, namely, particle swarm optimisation, to eliminate gradient-related problems. 

\vspace{-0.1cm}

This work aimed to improve the understanding of how PINNs performance is affected by  optimiser choice. Specifically, we considered the following algorithms spanning different optimiser categories: gradient descent (GD) without momentum, LBFGS \citep{Dennis1973ACO, Goodfellow-et-al-2016} (a second order quasi-Newton method), ADAM \citep{kingma2014adam} (an adaptive stochastic GD algorithm), and bouncing Born--Infeld (BBI) \citep{de2022born}. The first three optimisers are widely used in ML optimisation problems, whereas BBI is a recent realisation of a frictionless energy-conserving optimiser. See \cref{app:bbi} for details. 

\vspace{-0.1cm}
Moreover, we introduce a new method to study optimisation in neural networks, which provides training trajectory curvature data at low computational cost. Using this approach, we studied the evolution of different optimisers through the network parameter space during training. Further discussion is given in \cref{sec:curvature}. 

\vspace{-0.1cm}

%Furthermore, we changed the complexity of the loss function by varying the stiffness of the hyperbolic PDE.

%Due to the non-trivial  structure of the loss function bulk term, a simple PDE with hierarchical coefficients can exhibit a highly non-convex loss landscape.

Specifically, we apply PINNs to solve the linear advection equation  as described in \cref{sec:linear_advection}. Linear advection is a simple PDE with a single parameter, i.e., the wave speed $\beta$, whose variation can tune the complexity of the PINN landscape~\citep{Krishnapriyan2021}. Moreover, to understand how a network adjusts to different depths and widths, we considered two configurations of multi-layer perceptron architectures with different numbers of hidden layers and nodes per layer.

\vspace{-0.1cm}
The main contributions of this work are:
\begin{itemize}[leftmargin=.75cm]
    \item We found that the choice of optimisation algorithm significantly impacts the convergence of PINNs models. Specifically, second-order optimisers that do not directly follow gradient directions, such as LBFGS, performed better, as shown in \cref{fig:error_beta}.
    \item We introduce a new low-cost method for studying the dynamics of optimisers.
    This relies on defining a local reference frame and evaluating the local curvature of the training trajectory in the NN parameter space.   
    %This allowed us to reveal a non-trivial relation between the training trajectory local curvature in the parameter space and generalisation---measured by the convergence error on the entire domain.
    \item We found a negative correlation between PINNs convergence error and the newly introduced training trajectory curvature which is shown in \cref{fig:error_kappa}. This implies that good PINNs solutions lie in highly curved regions of the optimiser reference frame. Therefore, making PINNs converge is hard for those techniques designed to explore shallow landscapes that generalise well under perturbation. 
\vspace{-0.3cm}
\end{itemize}

\section{Background}
\vspace{-0.1cm}
\subsection{Tracking local curvature}\label{sec:curvature}

Solely observing the convergence performance of an optimiser will show which optimiser is the best for a given task but  will give limited information as to why. To move a step towards explaining why one optimiser is better than another, we analyze the trajectory of every optimiser in their local reference frame.% by introducing local trajectory curvatures.

To begin, let $\omegavec$ be the neural network parameters, the weight update rule for update $k$ is defined as $\Delta \boldsymbol{\omega}_k \equiv  \boldsymbol{\omega}_{k+1} - \boldsymbol{\omega}_k = \mathbf{V}(\boldsymbol{\omega}_k, \boldsymbol{\eta})$, where $\boldsymbol{\eta}$ are the model hyperparameters. Assuming the continuum time limit, the trajectory in the parameter space during training can be described by:
\begin{equation}\label{eq:V}
    \omegadot = \mathbf{V}(\boldsymbol{\omega}, \boldsymbol{\eta}),    
\end{equation}
where the overdot corresponds to a time derivative, i.e. $\omegadot = \mathrm{d} \omegavec/\mathrm{d}t$. When using gradient descent, \cref{eq:V} can be interpreted as the equation of motion of a classical particle in a friction-dominated setup and $\mathbf{V}(\boldsymbol{\omega}, \lambda) = -\lambda \frac{1}{n}\sum_i\nabla_{\boldsymbol{\omega}}\,\, \mathcal{L}(\mathbf{x}^{(i)}, \boldsymbol{\omega})$. Here, $\lambda$ is the learning rate, $n$ is the number of samples in the training set $\{\mathbf{x}^{(1)}, \mathbf{x}^{(2)}, \cdots, \mathbf{x}^{(n)}\}$, and $\mathcal{L}$ is the loss function.
For other optimisation algorithms, the definition of $\V$ changes and the ODE in \cref{eq:V} is no longer obtained via the Euler--Lagrange equations. Therefore, it does not have a dynamic interpretation. 
However, \cref{eq:V} can be efficiently used to understand the kinematics of the optimiser trajectories. This aspect is discussed further in \cref{sec:results}. 

To examine the motion in parameter space described by \cref{eq:V}, we introduce the unit vector tangential to the training trajectory, defined as
\begin{equation}
     \That = \frac{\omegadot}{\omegadotnorm}, \quad \textrm{with} \quad  \|\omegadot\|_2 = \sqrt{\langle\omegadot, \omegadot\rangle}.
\end{equation}
Here, $\That$ is a time-dependent vector in a $N_{\boldsymbol{w}}$-dimensional space, for a $N_{\boldsymbol{w}}$ parameter NN. To track the local training trajectory curvature in this time-dependent reference frame, we can define  the local (time-dependent) curvature, $\kappa_t$. This is the rate at which the tangent unit-vector changes with respect to time, and is given by:
\begin{equation}\label{eq:kappa}
    \kappa_t = \left\|\dx{\That}{t}\right\|_2 = \frac{1}{\omegadotnorm} \left[\omegaddot \cdot \omegaddot - \left(\dx{\omegadotnorm}{t}\right)^2\right]^{1/2}.
\end{equation}

Another parametrisation with a clearer geometric meaning is $\kappa_\omega=\left\|\dx{\That}{\omega}\right\|_2=\kappa_t/\|\omegadot\|_2$. This represents a local geometric quantity, i.e., the local curvature in the parameter space, removing the effects of time and trajectory speed\footnote{This is the local curvature of the training trajectory, not the curvature of the space of weights, which is flat.}. To calculate the curvature at each training step a first-order approximation is used, details of which are given in \cref{app:curvature_discretised}. 

In comparison with a Hessian-based curvature calculation, the method presented here has a significantly lower computational cost as it does not require the Hessian of the loss function. Furthermore, in most cases, an optimiser does not solely follow the gradient loss and it is generally non-trivial to find the function relating the loss, its derivatives, and the trajectory followed by the optimiser.
Therefore, although every optimiser is constructed with the aim of sharing its minima with the loss function, analyzing the loss Hessian may not characterise the optimiser trajectories and may not explain why a certain minimum is found. Moreover, different optimisers are based on different assumptions---such as energy conservation, friction-dominated motion, or others---further modifying the loss seen by the optimiser. \cref{fig:loss_trajectory} in \cref{app:curvature_discretised} gives a pictorial representation of the loss function (or its distorted version seen by the optimiser) and its relationship to the trajectory in the parameter space.

\vspace{-0.2cm}

\subsection{Linear Advection}\label{sec:linear_advection}
 The PDE used for the tests in this work was the one-dimensional linear advection equation on the periodic spatial domain $\Omega=[0,2\pi)$, given by
\begin{subequations}\label{eq:linadvec}
    \begin{align}
        \px{u}{t} + \beta \px{u}{x} &= 0, \quad \rm{for} \quad u:T\times\Omega\mapsto\mathbb{R}, \quad \Omega=[0,2 \pi), T \in [0,1], \\
        u(x,0) &= u_0(x), \\
        u(0,t) &= u(2 \pi, t), 
    \end{align}
\end{subequations}

where $\beta$ is the wave speed and $u_0(x)$ is the initial condition. The exact solution of this system can be straightforwardly found using the superposition of Bloch waves. To find approximate solutions to this system we applied PINNs with  loss functions similar to those used by \citet{Krishnapriyan2021}, defined as
\begin{subequations}
    \begin{align}
    \label{eq:loss}
        \mathcal{L}(\theta) &= \frac{1}{N_u} \sum_{i=1}^{N_u} \left(\hat{u} - u_0^i \right)^2  + \frac{1}{N_f} \sum_{i=1}^{N_f} \lambda_i \left(\px{\hat{u}}{t} + \beta \px{\hat{u}}{x} \right)^2 + \frac{1}{N_b} \sum_{i=1}^{N_b} \left(\hat{u}(\theta, 0, t) - \hat u(\theta, 2\pi, t)\right)^2.
    \end{align}
\end{subequations}
Here $\hat{u} = \hat{u}(\theta, x, t)$ is the neural network output, parameterised by $\theta$. The initial condition used through out this work was $u(x,0) = \sin(x)$ and $u(0,t)= u(2\pi,t)$. For all the experiments we fix $\lambda_i = 1$.

\section{Results and Discussion}\label{sec:results}
\begin{figure}
    \centering
    \subfloat[][]{\label{fig:error_beta}\adjustbox{width=0.42\linewidth, valign=b}{\begin{tikzpicture}
    \begin{axis}
    [
        axis line style={latex-latex},
        axis y line=left,
        axis x line=left,
        xmode=log,
        ymode=log,
        xlabel = {$\beta$},
        ylabel = {Median MSE},
        xmin = 1e0, xmax = 33,
        xtick = {1e0,1e1,1e2},
        ymin = 1e-7, ymax = 1,
        legend cell align={left},
        legend style={font=\scriptsize, at={(0.82, 0.03)},anchor=south west},
        %axis line style={draw=none},
        %tick style={draw=none},
        %x tick label style={/pgf/number format/.cd, fixed, fixed zerofill, precision=0, /tikz/.cd},
        %y tick label style={/pgf/number format/.cd, fixed, fixed zerofill, precision=1, /tikz/.cd},
    ]
        \addplot[line width=1pt, mark=*, mark size=1pt, color=Set1-A] table[x=beta, y=sgd0, col sep=comma]{./figs/data/beta.csv};
        \addlegendentry{GD};
        \addplot[line width=1pt, mark=*, mark size=1pt, color=Set1-A, dashed, forget plot] table[x=beta, y=sgd1, col sep=comma]{./figs/data/beta.csv};

        \addplot[line width=1pt, mark=*, mark size=1pt, color=Set1-B] table[x=beta, y=bbi0, col sep=comma]{./figs/data/beta.csv};
        \addlegendentry{BBI};
        \addplot[line width=1pt, mark=*, mark size=1pt, color=Set1-B, dashed, forget plot] table[x=beta, y=bbi1, col sep=comma]{./figs/data/beta.csv};

        \addplot[line width=1pt, mark=*, mark size=1pt, color=Set1-C] table[x=beta, y=adam0, col sep=comma]{./figs/data/beta.csv};
        \addlegendentry{ADAM};
        \addplot[line width=1pt, mark=*, mark size=1pt, color=Set1-C, dashed, forget plot] table[x=beta, y=adam1, col sep=comma]{./figs/data/beta.csv};

        \addplot[line width=1pt, mark=*, mark size=1pt, color=Set1-E] table[x=beta, y=lbfgs0, col sep=comma]{./figs/data/beta.csv};
        \addlegendentry{LBFGS};    
        \addplot[line width=1pt, mark=*, mark size=1pt, color=Set1-E, dashed, forget plot] table[x=beta, y=lbfgs1, col sep=comma]{./figs/data/beta.csv};

    \end{axis}
\end{tikzpicture}}}
    ~
    \subfloat[][]{\label{fig:error_kappa}\adjustbox{width=0.52\linewidth, valign=b}{\includegraphics{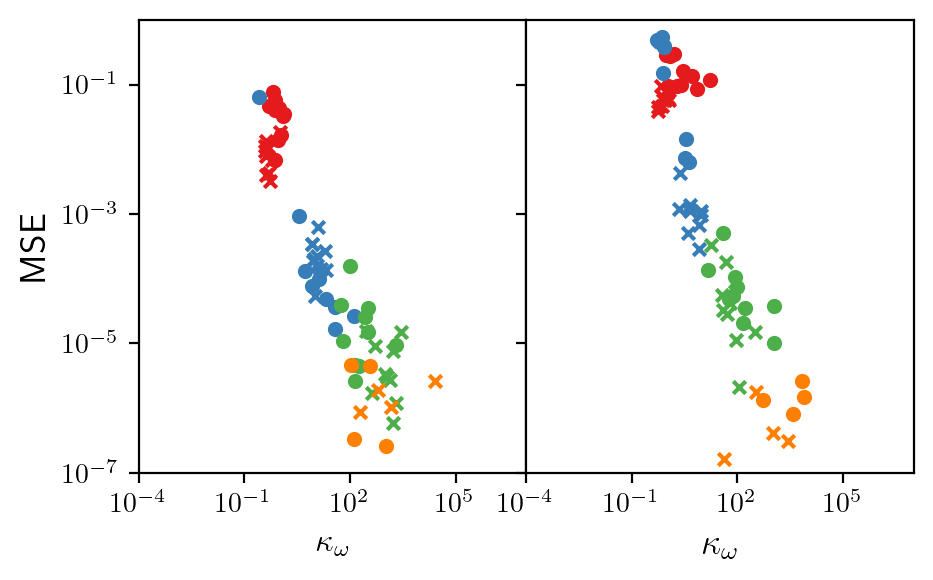}}}
    \caption{(a) Median MSE over 10 samples for various optimisers and $\beta$ values. \emph{Solid}/\emph{dashed} lines refer to small/large (S/L) network architectures. (b) Relation between final values of convergence (MSE)  and  curvature ($\kappa_\omega$) for $\beta=1$ (left) and $5$ (right). Dots and crosses refer to small (S) and large (L) networks, respectively. \label{fig:error}}
\vspace{-0.3cm}
\end{figure}
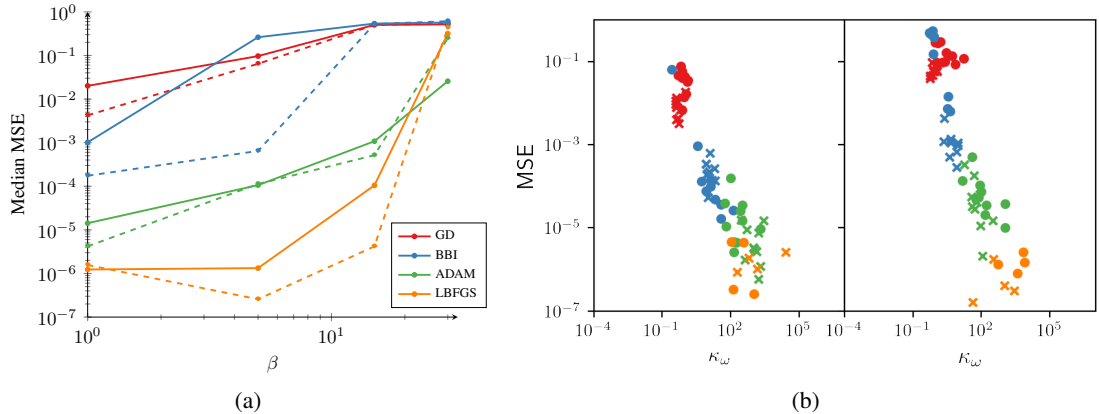

\cref{fig:error_beta} shows the median over 10 samples of the mean squared error (MSE) evaluated on the entire domain between the $\hat{u}$ predicted by the PINN and the analytical solution. Here the samples were formed by using 10 different random network initialisation. The corresponding training and test losses are shown in \cref{app:losses}. During testing, it was observed that the learning rate  significantly impacts the final performances; therefore, a grid search was performed over a range of learning rates to estimate the best configuration for each optimiser. See \cref{app:experiments} for the details. 
\vspace{-0.1cm}

Two multi-layer perception architectures were considered, with layer structures $S = [2, 25, 25, 1]$ and $L = [2, 50, 50, 50, 50, 1]$ resulting in \num{751} and \num{7851} trainable parameters respectively\footnote{These values keep the perturbative behaviour of the NN constant, i.e. depth/width constant~\citep{roberts_yaida_hanin_2022}.}. The number of training points was around 2000, as described in \cref{app:experiments}; therefore, the larger network is in the over-parameterised regime, i.e. it has more parameters than training data. \cref{fig:error_beta} shows that the large network generally has a lower error, in agreement with common knowledge in ML\footnote{From classical statistical learning, one expects that over-parameterised models over-fit. However, there is overwhelming evidence that large models generalise well in several cases of interest, e.g. \cite{Szegedy2015, huang2019gpipe}.}. This effect is most noticeable for BBI \footnote{The total phase space volume for BBI  can be analytically estimated and is parametrically large as the objective function is close to zero with an exponential dependence on the number of dimensions (see Appendix A.3 in \cite{de2022born}).}. A significant observation is that for all configurations of NN and optimiser, PINNs failed to produce good approximations for systems with large values of $\beta$.
\vspace{-0.1cm}

In \cref{fig:error_kappa}, we show the inverse relation  between the final values of MSE and the curvature $\kappa_\omega$. Here, only $\beta = 1,5$ were considered as they converged to reasonable errors for most of the optimisers, see \cref{app:training_dynamics}. Significantly, LBFGS achieves the highest values of $\kappa_\omega$ and the lowest error. Moreover, LBFGS has significant memory overhead but it typically converges in  fewer epochs compared to the other optimisers. Note that such a negative correlation in \cref{fig:error_kappa} links generalisation (measured by the MSE on the entire domain) with the curvature $\kappa_\omega$. Deriving their exact relation is non-trivial and we leave further investigations for future work. Nevertheless, we would like to stress that the high curvature values is not solely associated with orbiting the final local or global minima. In fact, this is due to $\kappa_\omega$ and MSE being negatively correlated throughout the whole trajectory, including its initial stages. We show examples of this behavior in \cref{app:training_dynamics}. 

%\begin{table}[ht]
%  \centering 
%  \begin{tabular}{M{5.1cm}M{5.1cm}}
%     $\qquad\qquad\qquad\qquad\beta = 1$     &       $\quad\beta = 5$ \\
%     \multicolumn{2}{r}{\includegraphics[width=10cm]{figures/Final_curvatures.png} }  
%     \vspace{-0.5cm}
%  \end{tabular}
%  \caption{\label{tab:MSEvskappa} Relation between final values of convergence (MSE of $u(x,t)$)  and intrinsic curvature ($\kappa_\omega$). S/L refer to small/large networks.}
%\end{table}
%}

% \vspace{-0.5cm}

\noindent
\textsc{Outlook.} In contrast to traditional ML tasks such as image recognition and NLP, ML methods applied to science require more accurate models which, in general, are trained with high-quality data. This suggests that the training process on data from physical phenomena can be fundamentally different from common ML applications. A promising future research direction would be exploring the connection between model generalisation and the flatness of minima for problems requiring high accuracy. For example, the work of \cite{dinh2017sharp, huang2020understanding} discuss this in the context of traditional ML tasks. Finally, in the future, it would be insightful to compare Hessian-based methods such as that of \cite{Michaud2023}, with our approach linking accuracy and local trajectory curvature.

\subsubsection*{Acknowledgments}
We thank Eloisa Bentivegna and Imran Nasim for  discussions. NF and WT acknowledge the UKRI support through the grant MR/T041862/1. The authors acknowledge the IBM Research Cognitive Computing Cluster service for providing resources that have contributed to the research results reported within this work.

\bibliography{reference}
\bibliographystyle{iclr2023_workshop}

\appendix
\section{Bouncing Born--Infeld (BBI) optimiser}\label{app:bbi}

Usual machine learning optimisation algorithms can be naturally described in physics terms as a particle moving down an irregular hill. Stochastic gradient descent with momentum is a standard example, as it can be viewed as a noisy and discretised version of a particle motion. Crucially, this is a friction-based evolution, such that the particle stops when there is insufficient kinetic energy to escape a minimum in the potential. \cite{de2022born} proposed an energy-conserving algorithm in which there is no friction and where the optimisation process slows down near the minima as this region dominates the phase space volume of the system. The algorithm of \cite{de2022born} is based on the relativistic Born--Infeld dynamics \citep{BornInfeld1934}, where the total potential energy ($V$) depends on the speed limit as $V=v_{\textrm{rel}}^2$, such that as $V \rightarrow 0$ the particle stops. For completeness, we summarise below BBI update rules \citep{de2022born}:
    \begin{subequations}
        \begin{align}
            \boldsymbol{\Pi}_{i+1}  &=   \boldsymbol{\Pi}_{i} - \frac{1}{2} \boldsymbol{\nabla} V_i \Delta t \left(\frac{V_i}{E} + \frac{E}{V_i}\right), \\
          \boldsymbol{\Theta}_{i+1} &=  \boldsymbol{\Theta}_{i} + \boldsymbol{\Pi}_{i+1} \Delta t \frac{V_i}{E},
         \end{align}
    \end{subequations}
with the nomenclature
\begin{itemize}[itemindent=1.2cm]
    \item[$\boldsymbol{\Theta}_{i}$] parameter vector with components $\theta_i$
    \item[$\boldsymbol{\Pi}_{i}$] momentum vector with components $\pi_i$
    \item[$i$] optimisation step number
    \item[$\Delta t$] optimisation step size (learning rate)
    \item[$V_i= V(\boldsymbol{\Theta}_i)$] the potential of the $i$-th step
    \item[$E = V_0 + \delta E$] constant dependent on the initialisation, and the additional initial energy parameter $\delta E$
    \item[$\boldsymbol{\Pi}_0$] the initial momentum, set as $ - \frac{\boldsymbol{\nabla} V(\boldsymbol{\Theta}_0)}{\mid\boldsymbol{\nabla} V(\boldsymbol{\Theta}_0)\mid} \sqrt{\frac{E^2}{V_0} -V_0}$.
\end{itemize}

As $E$ is constant, a particle can be trapped in long-lived orbits in motion. To  avoid such stable orbits and boost chaotic mixing, random bounces are introduced  by generating a new random momentum vectors with the same absolute momentum. There are three additional hyperparameters controlling bouncing: number of bounces ($N_b$), fixed timesteps for bounces ($T_0$), and progress-dependent timesteps for bounces ($T_1$).  Additionally momentum can be re-scaled to conserve energy lost  due to discretisation effects.
\section{Experimental Details}

\label{app:experiments}

\subsection{PINN training setup}

    \paragraph{Networks.} We considered both 2-hidden-layer and 4-hidden-layer fully-connect networks with 25 and 50 nodes per layer, respectively. We used hyperbolic tangents as activation functions in order to have a proper comparison with \cite{Krishnapriyan2021}.

    \paragraph{Data.} The training and test data are obtained by randomly sampling  $(x,t)$ points on the domain $\Omega=[0,2 \pi)$ and $T \in [0,1]$ using a grid of side  $n_x =256$  and $n_t=100$. We used $N_u = 100$ (initial conditions), $N_f =2000$ (bulk) and $N_b=80$ (periodic boundary conditions). Additionally, for the ADAM optimiser, we uniformly divide the total dataset into mini-batches of size $\mathcal{O}(400)$. For all cases, we split the data into training (80\%) and test (20\%) sets.

    \paragraph{Hyperparameters.} The learning rates and wave speeds ($\beta$) used for training are given in \cref{tab:lin_adv_scan}. For the specialised hyperparameters for ADAM ($\beta_1 = 0.9, \beta_2 = 0.999, \eps=10^{-8}$, weight-decay = 0), GD  (no mini-batches and no momentum) and LBFGS (max-iter = 20, tolerance-grad = $10^{-7}$, tolerance-change = $10^{-9}$, history-size = 100), we used the Pytorch default values.\footnote{\url{https://pytorch.org/docs/stable/optim.html}}  For BBI, we used $\Delta V =0$ (objective function shift), $\delta E =2$ (extra initial energy), $N_b =4$ (number of bounces), $T_0 =500$ (fixed timesteps for bounces), and $T_1=100$ (progress-dependent timesteps for bounces). We trained the models for 1000-5000 epochs depending on the convergence. Training dynamics (train and test losses) is shown in \cref{app:losses}.

\subsection{Learning Rate search for the Linear Advection}

Let us start by discussing the criteria  used to compare the optimisation algorithms.   Naturally, ADAM, BBI, GD and LBFGS  have different  hyperparameters. BBI, in particular, has specialised hyperparameters to boost chaotic mixing via random bounces (see \cref{app:bbi}). We note empirically that for the linear advection equation, the learning rate is the hyperparameter that impacts the most in the performances.
In Tab.\,\ref{tab:lin_adv_scan} 
 we show the results for the lowest average test losses using 5 trials with random initialised networks varying the learning rates in the range $[10^{-4}, 1].$ We trained the models for the first 300--1000 epochs depending on the convergence.  We note that  BBI and GD  are numerically unstable for learning rates above $\sim 0.1$.  By performing this search,  we  estimate the best scenario for each optimiser with respect to the learning rate. 
We acknowledge, however,  that significant changes in the other hyperparameters may impact the performances.

\begin{table}[tbhp]
    \captionsetup{position=top}
    \caption{\textbf{Learning rates for the lowest Test Loss.\label{tab:lin_adv_scan}}}
    \centering
    \subfloat[$\begin{bmatrix}2& 25& 25& 1\end{bmatrix}$]{\label{tab:network_1}
    \begin{tabular}{lrrrr}
        \toprule
        $\beta$ & BBI  & LBFGS  & GD  & Adam\\
        \midrule
        1  & 0.1    & 0.1      &0.01    & 0.001\\
        5  & 0.01    & 0.1      &0.01    & 0.001\\
        15 & 0.01   & 0.01      &0.0001  & 0.01\\
        30 & 0.01   & 0.01     &0.001   & 0.01\\
        \bottomrule
    \end{tabular}}
    ~
    \subfloat[$\begin{bmatrix}2& 50& 50& 50& 50 &1\end{bmatrix}$]{\label{tab:network_2}
    \begin{tabular}{lrrrr}
        \toprule
        $\beta$ & BBI  & LBFGS   & GD  & Adam\\
        \midrule
        1   &    0.01   & 0.1    &   0.01 & 0.0001\\
        5   &    0.01     & 0.1    &   0.01 & 0.001\\
        15  &    0.01     & 1    &   0.01  & 0.001\\
        30  &    0.01     & 0.001    &   0.001  & 0.001\\
        \bottomrule
    \end{tabular}}
    \captionsetup{position=bottom}
\end{table}

\section{Further details on the Curvature}\label{app:curvature_discretised}

\subsection{Curvature discretisation}
    To derive a discretised version of the curvature with respect to the training steps, consider the first-order approximation of \cref{eq:V} given by
    \begin{equation}
           \omegadot_{k + 1} \approx \omegavec_{k + 1} - \omegavec_k = \V_k.
    \end{equation}
    To obtain an approximation for the curvature in \cref{eq:kappa}, the following steps can be taken
    \begin{subequations}
        \begin{align}
            \omegaddot_{k + 1} &\approx \V_k - \V_{k-1}, \\
            |\omegadot_{k + 1}|  &\approx \sqrt{\langle\V_k, \V_k\rangle}, \\
             \frac{d}{dt}|\omegadot_{k + 1}| &\approx \frac{\langle\V_k, \V_k\rangle - \langle\V_k,\V_{k-1}\rangle}{\sqrt{\langle\V_{k},\V_k\rangle}}, \\
            \kappa_{t,k+1}
            & \approx \frac{1}{\sqrt{\langle\V_k,\V_k\rangle}} \left[\langle \V_{k-1},\V_{k-1}\rangle - \frac{\langle\V_{k-1},\V_k\rangle^2}{\langle\V_k,\V_k\rangle}\right]^{1/2}.
         \end{align}
    \end{subequations}

\subsection{Local curvature and the loss function}

Here we comment on the relation between $\kappa_\omega$, i.e. the time-independent local curvature of the training trajectory in the parameter space, and the loss function. Following a physics intuition, given an objective function, $\mathcal{L}(\boldsymbol{\omega})$, we tend to see it as a potential and assume that our trajectory lies on the hypersurface described by $\mathcal{L}(\boldsymbol{\omega})$. This is not true, as different optimisers follow different theories that modify the loss landscape and the regime (e.g. energy-conserving and friction-dominated motion). Nevertheless, given the form of \cref{eq:V}, we can reinterpret the optimiser as if it were a friction-dominated classical algorithm, such as GD. This gives us a hint about how large a gradient should be to induce that step in the GD algorithm, and it tells us how curved the distorted loss landscape, seen by the optimiser, is.

\begin{figure}[tbhp]
    \centering
    \includegraphics[width=0.5\textwidth]{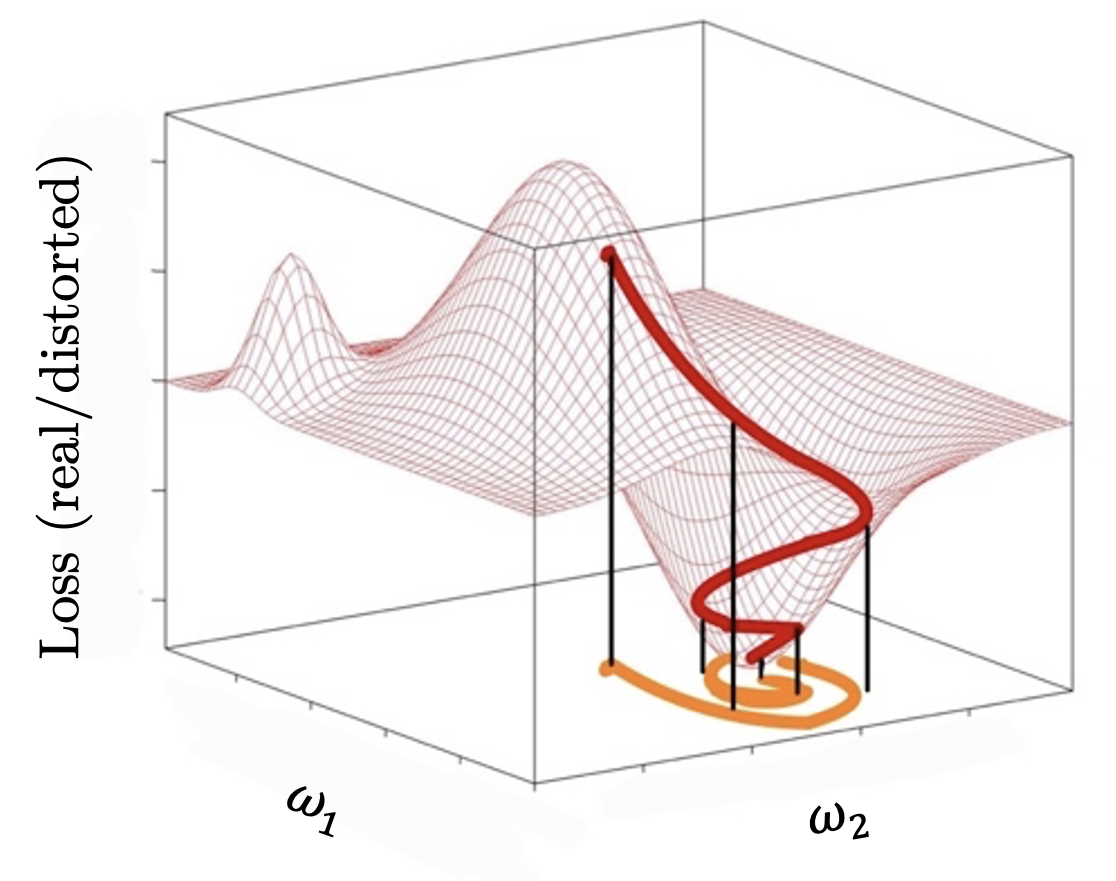}
    \caption{\label{fig:loss_trajectory}Pictorial view of the relation between the loss function (or its distorted version seen by the optimiser). The orange curve on the plane $\omega_1-\omega_2$ represents the trajectory followed by the optimiser on the parameter space.}
\end{figure}

\cref{fig:loss_trajectory} shows a pictorial view of the relation between the loss function, or its distorted version seen by the optimiser, and the trajectory in the parameter space. The path followed on the loss hypersurface is depicted with a red curve, while its projection on the parameter space is in orange. The orange curve is precisely the focus of this study. Indeed, comparing trajectories living on different loss---or distorted loss---hypersurfaces is an ill-defined problem. On the other hand, monitoring the projected trajectories and their local curvature, $\kappa_\omega$, tells us how curved is the fictitious loss landscape seen by each optimiser. 

\section{Additional Plots}
\label{app:plots}
\subsection{Training and Test Losses}
\label{app:losses}
\begin{figure}[tbhp]
    \centering
    \subfloat[$\beta=1$, Network 1]{\adjustbox{width=0.42\linewidth,valign=b}{\begin{tikzpicture}
    \begin{axis}
    [
        axis line style={latex-latex},
        axis y line=left,
        axis x line=left,
        width=9cm,
        height=6cm,
        xmode=linear,
        ymode=log,
        xlabel = {Epoch},
        ylabel = {Loss},
        xmin = 0, xmax = 5200,
        xtick = {0,1000,2000,3000,4000,5000},
        ymin = 1e-6, ymax = 1,
        ytick = {1e0,1e-1,1e-2,1e-3,1e-4,1e-5,1e-6,1e-7},
        legend cell align={left},
        legend style={font=\small, at={(0.97, 0.97)},anchor=north east},
        %axis line style={draw=none},
        %tick style={draw=none},
        %x tick label style={/pgf/number format/.cd, fixed, fixed zerofill, precision=0, /tikz/.cd},
        %y tick label style={/pgf/number format/.cd, fixed, fixed zerofill, precision=1, /tikz/.cd},
    ]
        
        \addplot[line width=1pt, color=Set1-A] table[x=epoch, y=train_loss_total, col sep=comma]{./figs/data/Adam0.001_beta1_NN0.csv};
        
        \addplot[line width=1pt, color=Set1-B] table[x=epoch, y=bc_loss_train, col sep=comma]{./figs/data/Adam0.001_beta1_NN0.csv};
        
        \addplot[line width=1pt, color=Set1-C] table[x=epoch, y=bulk_loss_train, col sep=comma]{./figs/data/Adam0.001_beta1_NN0.csv};

        \addplot[line width=1pt, color=Set1-D] table[x=epoch, y=bcp_loss_train, col sep=comma]{./figs/data/Adam0.001_beta1_NN0.csv};

        \addplot[line width=1pt, color=Set1-A, forget plot, densely dotted] table[x=epoch, y=test_loss_total, col sep=comma]{./figs/data/Adam0.001_beta1_NN0.csv};
        
        \addplot[line width=1pt, color=Set1-B, forget plot, densely dotted] table[x=epoch, y=bc_loss_test, col sep=comma]{./figs/data/Adam0.001_beta1_NN0.csv};
        
        \addplot[line width=1pt, color=Set1-C, forget plot,  densely dotted] table[x=epoch, y=bulk_loss_test, col sep=comma]{./figs/data/Adam0.001_beta1_NN0.csv};
        
        \addplot[line width=1pt, color=Set1-D, forget plot, densely dotted] table[x=epoch, y=bcp_loss_test, col sep=comma]{./figs/data/Adam0.001_beta1_NN0.csv};
        
        \addlegendentry{Total};
        \addlegendentry{IC};
        \addlegendentry{Bulk};
        \addlegendentry{BC};

    \end{axis}
\end{tikzpicture}}}
    \subfloat[$\beta=1$, Network 2]{\adjustbox{width=0.42\linewidth,valign=b}{\begin{tikzpicture}
    \begin{axis}
    [
        axis line style={latex-latex},
        axis y line=left,
        axis x line=left,
        width=9cm,
        height=6cm,
        xmode=linear,
        ymode=log,
        xlabel = {Epoch},
        ylabel = {Loss},
        xmin = 0, xmax = 5200,
        xtick = {0,1000,2000,3000,4000,5000},
        ymin = 1e-6, ymax = 1,
        ytick = {1e0,1e-1,1e-2,1e-3,1e-4,1e-5,1e-6,1e-7},
        legend cell align={left},
        legend style={font=\scriptsize, at={(0.97, 0.97)},anchor=north east},
        %axis line style={draw=none},
        %tick style={draw=none},
        %x tick label style={/pgf/number format/.cd, fixed, fixed zerofill, precision=0, /tikz/.cd},
        %y tick label style={/pgf/number format/.cd, fixed, fixed zerofill, precision=1, /tikz/.cd},
    ]
        
        \addplot[line width=1pt, color=Set1-A] table[x=epoch, y=train_loss_total, col sep=comma]{./figs/data/Adam0.0001_beta1_NN1.csv};
        
        \addplot[line width=1pt, color=Set1-B] table[x=epoch, y=bc_loss_train, col sep=comma]{./figs/data/Adam0.0001_beta1_NN1.csv};
        
        \addplot[line width=1pt, color=Set1-C] table[x=epoch, y=bulk_loss_train, col sep=comma]{./figs/data/Adam0.0001_beta1_NN1.csv};
        
        \addplot[line width=1pt, color=Set1-D] table[x=epoch, y=bcp_loss_train, col sep=comma]{./figs/data/Adam0.0001_beta1_NN1.csv};

        \addplot[line width=1pt, color=Set1-A, densely dotted] table[x=epoch, y=test_loss_total, col sep=comma]{./figs/data/Adam0.0001_beta1_NN1.csv};
        
        \addplot[line width=1pt, color=Set1-B, densely dotted] table[x=epoch, y=bc_loss_test, col sep=comma]{./figs/data/Adam0.0001_beta1_NN1.csv};
        
        \addplot[line width=1pt, color=Set1-C, densely dotted] table[x=epoch, y=bulk_loss_test, col sep=comma]{./figs/data/Adam0.0001_beta1_NN1.csv};
        
        \addplot[line width=1pt, color=Set1-D, densely dotted] table[x=epoch, y=bcp_loss_test, col sep=comma]{./figs/data/Adam0.0001_beta1_NN1.csv};

    \end{axis}
\end{tikzpicture}}}
    \\
    \subfloat[$\beta=5$, Network 1]{\adjustbox{width=0.42\linewidth,valign=b}{\begin{tikzpicture}
    \begin{axis}
    [
        axis line style={latex-latex},
        axis y line=left,
        axis x line=left,
        width=9cm,
        height=6cm,
        xmode=linear,
        ymode=log,
        xlabel = {Epoch},
        ylabel = {Loss},
        xmin = 0, xmax = 5200,
        xtick = {0,1000,2000,3000,4000,5000},
        ymin = 1e-6, ymax = 1,
        ytick = {1e0,1e-1,1e-2,1e-3,1e-4,1e-5,1e-6,1e-7},
        legend cell align={left},
        legend style={font=\scriptsize, at={(0.97, 0.97)},anchor=north east},
        %axis line style={draw=none},
        %tick style={draw=none},
        %x tick label style={/pgf/number format/.cd, fixed, fixed zerofill, precision=0, /tikz/.cd},
        %y tick label style={/pgf/number format/.cd, fixed, fixed zerofill, precision=1, /tikz/.cd},
    ]
        
        \addplot[line width=1pt, color=Set1-A] table[x=epoch, y=train_loss_total, col sep=comma]{./figs/data/Adam0.001_beta5_NN0.csv};
        
        \addplot[line width=1pt, color=Set1-B] table[x=epoch, y=bc_loss_train, col sep=comma]{./figs/data/Adam0.001_beta5_NN0.csv};
        
        \addplot[line width=1pt, color=Set1-C] table[x=epoch, y=bulk_loss_train, col sep=comma]{./figs/data/Adam0.001_beta5_NN0.csv};
        
        \addplot[line width=1pt, color=Set1-D] table[x=epoch, y=bcp_loss_train, col sep=comma]{./figs/data/Adam0.001_beta5_NN0.csv};

        \addplot[line width=1pt, color=Set1-A, forget plot, densely dashed] table[x=epoch, y=test_loss_total, col sep=comma]{./figs/data/Adam0.001_beta5_NN0.csv};
        
        \addplot[line width=1pt, color=Set1-B, forget plot, densely dashed] table[x=epoch, y=bc_loss_test, col sep=comma]{./figs/data/Adam0.001_beta5_NN0.csv};
        
        \addplot[line width=1pt, color=Set1-C, forget plot,  densely dotted] table[x=epoch, y=bulk_loss_test, col sep=comma]{./figs/data/Adam0.001_beta5_NN0.csv};
        
        \addplot[line width=1pt, color=Set1-D, forget plot, densely dotted] table[x=epoch, y=bcp_loss_test, col sep=comma]{./figs/data/Adam0.001_beta5_NN0.csv};

    \end{axis}
\end{tikzpicture}}}
    \subfloat[$\beta=5$, Network 2]{\adjustbox{width=0.42\linewidth,valign=b}{\begin{tikzpicture}
    \begin{axis}
    [
        axis line style={latex-latex},
        axis y line=left,
        axis x line=left,
        width=9cm,
        height=6cm,
        xmode=linear,
        ymode=log,
        xlabel = {Epoch},
        ylabel = {Loss},
        xmin = 0, xmax = 5200,
        xtick = {0,1000,2000,3000,4000,5000},
        ymin = 1e-6, ymax = 1,
        ytick = {1e0,1e-1,1e-2,1e-3,1e-4,1e-5,1e-6,1e-7},
        legend cell align={left},
        legend style={font=\scriptsize, at={(0.97, 0.97)},anchor=north east},
        %axis line style={draw=none},
        %tick style={draw=none},
        %x tick label style={/pgf/number format/.cd, fixed, fixed zerofill, precision=0, /tikz/.cd},
        %y tick label style={/pgf/number format/.cd, fixed, fixed zerofill, precision=1, /tikz/.cd},
    ]
        
        \addplot[line width=1pt, color=Set1-A] table[x=epoch, y=train_loss_total, col sep=comma]{./figs/data/Adam0.001_beta5_NN1.csv};
        
        \addplot[line width=1pt, color=Set1-B] table[x=epoch, y=bc_loss_train, col sep=comma]{./figs/data/Adam0.001_beta5_NN1.csv};
        
        \addplot[line width=1pt, color=Set1-C] table[x=epoch, y=bulk_loss_train, col sep=comma]{./figs/data/Adam0.001_beta5_NN1.csv};
        
        \addplot[line width=1pt, color=Set1-D] table[x=epoch, y=bcp_loss_train, col sep=comma]{./figs/data/Adam0.001_beta5_NN1.csv};

        \addplot[line width=1pt, color=Set1-A, densely dotted] table[x=epoch, y=test_loss_total, col sep=comma]{./figs/data/Adam0.001_beta5_NN1.csv};
        
        \addplot[line width=1pt, color=Set1-B, densely dotted] table[x=epoch, y=bc_loss_test, col sep=comma]{./figs/data/Adam0.001_beta5_NN1.csv};
        
        \addplot[line width=1pt, color=Set1-C, densely dotted] table[x=epoch, y=bulk_loss_test, col sep=comma]{./figs/data/Adam0.001_beta5_NN1.csv};
        
        \addplot[line width=1pt, color=Set1-D, densely dotted] table[x=epoch, y=bcp_loss_test, col sep=comma]{./figs/data/Adam0.001_beta5_NN1.csv};

    \end{axis}
\end{tikzpicture}}}
    \\
    \subfloat[$\beta=15$, Network 1]{\adjustbox{width=0.42\linewidth,valign=b}{\begin{tikzpicture}
    \begin{axis}
    [
        axis line style={latex-latex},
        axis y line=left,
        axis x line=left,
        width=9cm,
        height=6cm,
        xmode=linear,
        ymode=log,
        xlabel = {Epoch},
        ylabel = {Loss},
        xmin = 0, xmax = 5200,
        xtick = {0,1000,2000,3000,4000,5000},
        ymin = 1e-6, ymax = 1,
        ytick = {1e0,1e-1,1e-2,1e-3,1e-4,1e-5,1e-6,1e-7},
        legend cell align={left},
        legend style={font=\scriptsize, at={(0.97, 0.97)},anchor=north east},
        %axis line style={draw=none},
        %tick style={draw=none},
        %x tick label style={/pgf/number format/.cd, fixed, fixed zerofill, precision=0, /tikz/.cd},
        %y tick label style={/pgf/number format/.cd, fixed, fixed zerofill, precision=1, /tikz/.cd},
    ]
        
        \addplot[line width=1pt, color=Set1-A] table[x=epoch, y=train_loss_total, col sep=comma]{./figs/data/Adam0.01_beta15_NN0.csv};
        
        \addplot[line width=1pt, color=Set1-B] table[x=epoch, y=bc_loss_train, col sep=comma]{./figs/data/Adam0.01_beta15_NN0.csv};
        
        \addplot[line width=1pt, color=Set1-C] table[x=epoch, y=bulk_loss_train, col sep=comma]{./figs/data/Adam0.01_beta15_NN0.csv};
        
        \addplot[line width=1pt, color=Set1-D] table[x=epoch, y=bcp_loss_train, col sep=comma]{./figs/data/Adam0.01_beta15_NN0.csv};

        \addplot[line width=1pt, color=Set1-A, densely dotted] table[x=epoch, y=test_loss_total, col sep=comma]{./figs/data/Adam0.01_beta15_NN0.csv};
        
        \addplot[line width=1pt, color=Set1-B, densely dotted] table[x=epoch, y=bc_loss_test, col sep=comma]{./figs/data/Adam0.01_beta15_NN0.csv};
        
        \addplot[line width=1pt, color=Set1-C,  densely dotted] table[x=epoch, y=bulk_loss_test, col sep=comma]{./figs/data/Adam0.01_beta15_NN0.csv};
        
        \addplot[line width=1pt, color=Set1-D, densely dotted] table[x=epoch, y=bcp_loss_test, col sep=comma]{./figs/data/Adam0.01_beta15_NN0.csv};

    \end{axis}
\end{tikzpicture}}}
    \subfloat[$\beta=15$, Network 2]{\adjustbox{width=0.42\linewidth,valign=b}{\begin{tikzpicture}
    \begin{axis}
    [
        axis line style={latex-latex},
        axis y line=left,
        axis x line=left,
        width=9cm,
        height=6cm,
        xmode=linear,
        ymode=log,
        xlabel = {Epoch},
        ylabel = {Loss},
        xmin = 0, xmax = 5200,
        xtick = {0,1000,2000,3000,4000,5000},
        ymin = 1e-6, ymax = 1,
        ytick = {1e0,1e-1,1e-2,1e-3,1e-4,1e-5,1e-6,1e-7},
        legend cell align={left},
        legend style={font=\scriptsize, at={(0.97, 0.97)},anchor=north east},
        %axis line style={draw=none},
        %tick style={draw=none},
        %x tick label style={/pgf/number format/.cd, fixed, fixed zerofill, precision=0, /tikz/.cd},
        %y tick label style={/pgf/number format/.cd, fixed, fixed zerofill, precision=1, /tikz/.cd},
    ]
        
        \addplot[line width=1pt, color=Set1-A] table[x=epoch, y=train_loss_total, col sep=comma]{./figs/data/Adam0.001_beta15_NN1.csv};
        
        \addplot[line width=1pt, color=Set1-B] table[x=epoch, y=bc_loss_train, col sep=comma]{./figs/data/Adam0.001_beta15_NN1.csv};
        
        \addplot[line width=1pt, color=Set1-C] table[x=epoch, y=bulk_loss_train, col sep=comma]{./figs/data/Adam0.001_beta15_NN1.csv};
        
        \addplot[line width=1pt, color=Set1-D] table[x=epoch, y=bcp_loss_train, col sep=comma]{./figs/data/Adam0.001_beta15_NN1.csv};

        \addplot[line width=1pt, color=Set1-A, densely dotted] table[x=epoch, y=test_loss_total, col sep=comma]{./figs/data/Adam0.001_beta15_NN1.csv};
        
        \addplot[line width=1pt, color=Set1-B, densely dotted] table[x=epoch, y=bc_loss_test, col sep=comma]{./figs/data/Adam0.001_beta15_NN1.csv};
        
        \addplot[line width=1pt, color=Set1-C, densely dotted] table[x=epoch, y=bulk_loss_test, col sep=comma]{./figs/data/Adam0.001_beta15_NN1.csv};
        
        \addplot[line width=1pt, color=Set1-D, densely dotted] table[x=epoch, y=bcp_loss_test, col sep=comma]{./figs/data/Adam0.001_beta15_NN1.csv};

    \end{axis}
\end{tikzpicture}}}
    \\
    \subfloat[$\beta=30$, Network 1]{\adjustbox{width=0.42\linewidth,valign=b}{\begin{tikzpicture}
    \begin{axis}
    [
        axis line style={latex-latex},
        axis y line=left,
        axis x line=left,
        width=9cm,
        height=6cm,
        xmode=linear,
        ymode=log,
        xlabel = {Epoch},
        ylabel = {Loss},
        xmin = 0, xmax = 5200,
        xtick = {0,1000,2000,3000,4000,5000},
        ymin = 1e-6, ymax = 1,
        ytick = {1e0,1e-1,1e-2,1e-3,1e-4,1e-5,1e-6,1e-7},
        legend cell align={left},
        legend style={font=\scriptsize, at={(0.97, 0.97)},anchor=north east},
        %axis line style={draw=none},
        %tick style={draw=none},
        %x tick label style={/pgf/number format/.cd, fixed, fixed zerofill, precision=0, /tikz/.cd},
        %y tick label style={/pgf/number format/.cd, fixed, fixed zerofill, precision=1, /tikz/.cd},
    ]
        
        \addplot[line width=1pt, color=Set1-A] table[x=epoch, y=train_loss_total, col sep=comma]{./figs/data/Adam0.01_beta30_NN0.csv};
        
        \addplot[line width=1pt, color=Set1-B] table[x=epoch, y=bc_loss_train, col sep=comma]{./figs/data/Adam0.01_beta30_NN0.csv};
        
        \addplot[line width=1pt, color=Set1-C] table[x=epoch, y=bulk_loss_train, col sep=comma]{./figs/data/Adam0.01_beta30_NN0.csv};
        
        \addplot[line width=1pt, color=Set1-D] table[x=epoch, y=bcp_loss_train, col sep=comma]{./figs/data/Adam0.01_beta30_NN0.csv};

        \addplot[line width=1pt, color=Set1-A, densely dotted] table[x=epoch, y=test_loss_total, col sep=comma]{./figs/data/Adam0.01_beta30_NN0.csv};
        
        \addplot[line width=1pt, color=Set1-B, densely dotted] table[x=epoch, y=bc_loss_test, col sep=comma]{./figs/data/Adam0.01_beta30_NN0.csv};
        
        \addplot[line width=1pt, color=Set1-C, densely dotted] table[x=epoch, y=bulk_loss_test, col sep=comma]{./figs/data/Adam0.01_beta30_NN0.csv};
        
        \addplot[line width=1pt, color=Set1-D, densely dotted] table[x=epoch, y=bcp_loss_test, col sep=comma]{./figs/data/Adam0.01_beta30_NN0.csv};

    \end{axis}
\end{tikzpicture}}}
    \subfloat[$\beta=30$, Network 2]{\adjustbox{width=0.42\linewidth,valign=b}{\begin{tikzpicture}
    \begin{axis}
    [
        axis line style={latex-latex},
        axis y line=left,
        axis x line=left,
        width=9cm,
        height=6cm,
        xmode=linear,
        ymode=log,
        xlabel = {Epoch},
        ylabel = {Loss},
        xmin = 0, xmax = 5200,
        xtick = {0,1000,2000,3000,4000,5000},
        ymin = 1e-6, ymax = 1,
        ytick = {1e0,1e-1,1e-2,1e-3,1e-4,1e-5,1e-6,1e-7},
        legend cell align={left},
        legend style={font=\scriptsize, at={(0.97, 0.97)},anchor=north east},
        %axis line style={draw=none},
        %tick style={draw=none},
        %x tick label style={/pgf/number format/.cd, fixed, fixed zerofill, precision=0, /tikz/.cd},
        %y tick label style={/pgf/number format/.cd, fixed, fixed zerofill, precision=1, /tikz/.cd},
    ]
        
        \addplot[line width=1pt, color=Set1-A] table[x=epoch, y=train_loss_total, col sep=comma]{./figs/data/Adam0.001_beta30_NN1.csv};
        
        \addplot[line width=1pt, color=Set1-B] table[x=epoch, y=bc_loss_train, col sep=comma]{./figs/data/Adam0.001_beta30_NN1.csv};
        
        \addplot[line width=1pt, color=Set1-C] table[x=epoch, y=bulk_loss_train, col sep=comma]{./figs/data/Adam0.001_beta30_NN1.csv};
        
        \addplot[line width=1pt, color=Set1-D] table[x=epoch, y=bcp_loss_train, col sep=comma]{./figs/data/Adam0.001_beta30_NN1.csv};

        \addplot[line width=1pt, color=Set1-A, densely dotted] table[x=epoch, y=test_loss_total, col sep=comma]{./figs/data/Adam0.001_beta30_NN1.csv};
        
        \addplot[line width=1pt, color=Set1-B, densely dotted] table[x=epoch, y=bc_loss_test, col sep=comma]{./figs/data/Adam0.001_beta30_NN1.csv};
        
        \addplot[line width=1pt, color=Set1-C, densely dotted] table[x=epoch, y=bulk_loss_test, col sep=comma]{./figs/data/Adam0.001_beta30_NN1.csv};
        
        \addplot[line width=1pt, color=Set1-D, densely dotted] table[x=epoch, y=bcp_loss_test, col sep=comma]{./figs/data/Adam0.001_beta30_NN1.csv};

    \end{axis}
\end{tikzpicture}}}
    \caption{\label{fig:adam_loss}Median loss versus epoch over 10 samples when using ADAM. \emph{Solid} line is the total, and the other lines show the separate contributions. \emph{dashed} for initial condition, \emph{dotted} for the bulk, and \emph{dash-dotted} for the periodic boundary.}
\end{figure}
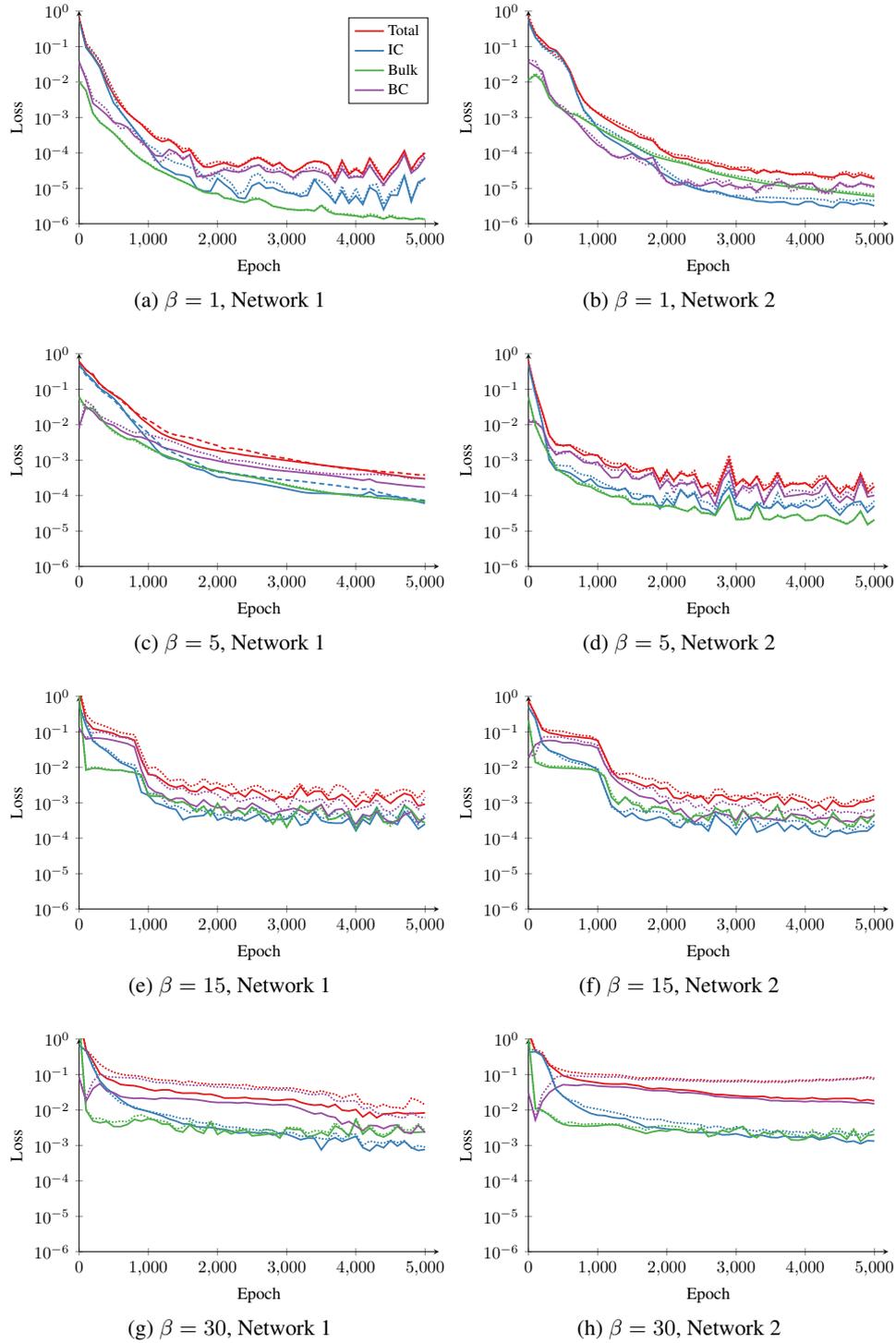

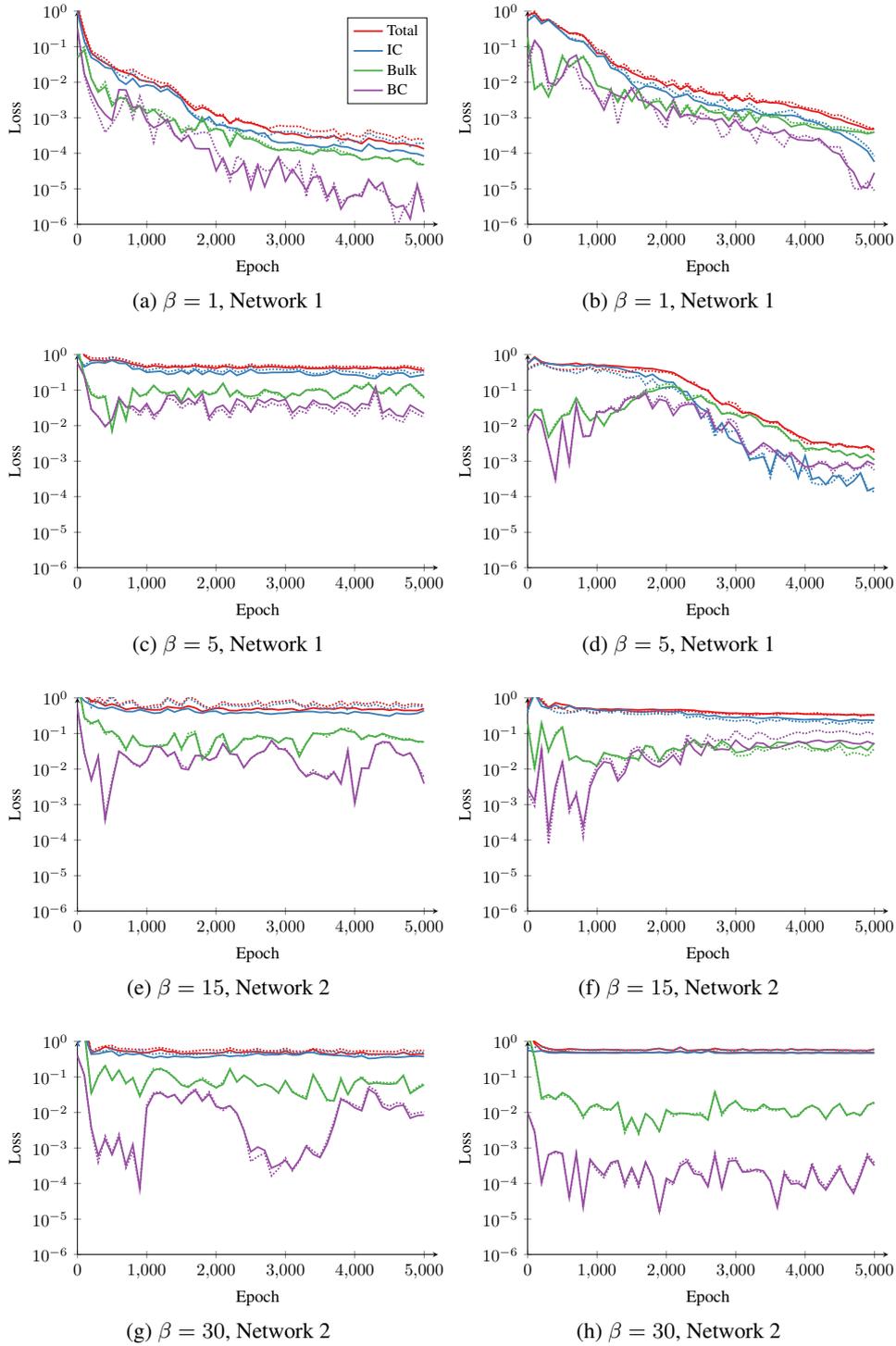
\begin{figure}[tbhp]
    \centering
    \subfloat[$\beta=1$, Network 1]{\adjustbox{width=0.42\linewidth,valign=b}{\begin{tikzpicture}
    \begin{axis}
    [
        axis line style={latex-latex},
        axis y line=left,
        axis x line=left,
        width=9cm,
        height=6cm,
        xmode=linear,
        ymode=log,
        xlabel = {Epoch},
        ylabel = {Loss},
        xmin = 0, xmax = 5200,
        xtick = {0,1000,2000,3000,4000,5000},
        ymin = 1e-6, ymax = 1,
        ytick = {1e0,1e-1,1e-2,1e-3,1e-4,1e-5,1e-6,1e-7},
        legend cell align={left},
        legend style={font=\small, at={(0.97, 0.97)},anchor=north east},
        %axis line style={draw=none},
        %tick style={draw=none},
        %x tick label style={/pgf/number format/.cd, fixed, fixed zerofill, precision=0, /tikz/.cd},
        %y tick label style={/pgf/number format/.cd, fixed, fixed zerofill, precision=1, /tikz/.cd},
    ]
        
        \addplot[line width=1pt, color=Set1-A] table[x=epoch, y=train_loss_total, col sep=comma]{./figs/data/BBI0.1_beta1_NN0.csv};
        \addplot[line width=1pt, color=Set1-B] table[x=epoch, y=bc_loss_train, col sep=comma]{./figs/data/BBI0.1_beta1_NN0.csv};
        \addplot[line width=1pt, color=Set1-C] table[x=epoch, y=bulk_loss_train, col sep=comma]{./figs/data/BBI0.1_beta1_NN0.csv};
        \addplot[line width=1pt, color=Set1-D] table[x=epoch, y=bcp_loss_train, col sep=comma]{./figs/data/BBI0.1_beta1_NN0.csv};
        
        \addplot[line width=1pt, color=Set1-A, densely dotted] table[x=epoch, y=test_loss_total, col sep=comma]{./figs/data/BBI0.1_beta1_NN0.csv};
        \addplot[line width=1pt, color=Set1-B, densely dotted] table[x=epoch, y=bc_loss_test, col sep=comma]{./figs/data/BBI0.1_beta1_NN0.csv};
        \addplot[line width=1pt, color=Set1-C, densely dotted] table[x=epoch, y=bulk_loss_test, col sep=comma]{./figs/data/BBI0.1_beta1_NN0.csv};
        \addplot[line width=1pt, color=Set1-D, densely dotted] table[x=epoch, y=bcp_loss_test, col sep=comma]{./figs/data/BBI0.1_beta1_NN0.csv};
        
        \addlegendentry{Total};
        \addlegendentry{IC};
        \addlegendentry{Bulk};
        \addlegendentry{BC};

    \end{axis}
\end{tikzpicture}}}
    \subfloat[$\beta=1$, Network 1]{\adjustbox{width=0.42\linewidth,valign=b}{\begin{tikzpicture}
    \begin{axis}
    [
        axis line style={latex-latex},
        axis y line=left,
        axis x line=left,
        width=9cm,
        height=6cm,
        xmode=linear,
        ymode=log,
        xlabel = {Epoch},
        ylabel = {Loss},
        xmin = 0, xmax = 5200,
        xtick = {0,1000,2000,3000,4000,5000},
        ymin = 1e-6, ymax = 1,
        ytick = {1e0,1e-1,1e-2,1e-3,1e-4,1e-5,1e-6,1e-7},
        legend cell align={left},
        legend style={font=\scriptsize, at={(0.97, 0.97)},anchor=north east},
        %axis line style={draw=none},
        %tick style={draw=none},
        %x tick label style={/pgf/number format/.cd, fixed, fixed zerofill, precision=0, /tikz/.cd},
        %y tick label style={/pgf/number format/.cd, fixed, fixed zerofill, precision=1, /tikz/.cd},
    ]
        
        \addplot[line width=1pt, color=Set1-A] table[x=epoch, y=train_loss_total, col sep=comma]{./figs/data/BBI0.01_beta1_NN1.csv};
        
        \addplot[line width=1pt, color=Set1-B] table[x=epoch, y=bc_loss_train, col sep=comma]{./figs/data/BBI0.01_beta1_NN1.csv};
        \addplot[line width=1pt, color=Set1-C] table[x=epoch, y=bulk_loss_train, col sep=comma]{./figs/data/BBI0.01_beta1_NN1.csv};
        \addplot[line width=1pt, color=Set1-D] table[x=epoch, y=bcp_loss_train, col sep=comma]{./figs/data/BBI0.01_beta1_NN1.csv};

        \addplot[line width=1pt, color=Set1-A, densely dotted] table[x=epoch, y=test_loss_total, col sep=comma]{./figs/data/BBI0.01_beta1_NN1.csv};
        
        \addplot[line width=1pt, color=Set1-B, densely dotted] table[x=epoch, y=bc_loss_test, col sep=comma]{./figs/data/BBI0.01_beta1_NN1.csv};
        \addplot[line width=1pt, color=Set1-C, densely dotted] table[x=epoch, y=bulk_loss_test, col sep=comma]{./figs/data/BBI0.01_beta1_NN1.csv};
        \addplot[line width=1pt, color=Set1-D, densely dotted] table[x=epoch, y=bcp_loss_test, col sep=comma]{./figs/data/BBI0.01_beta1_NN1.csv};

    \end{axis}
\end{tikzpicture}}}
    \\
    \subfloat[$\beta=5$, Network 1]{\adjustbox{width=0.42\linewidth,valign=b}{\begin{tikzpicture}
    \begin{axis}
    [
        axis line style={latex-latex},
        axis y line=left,
        axis x line=left,
        width=9cm,
        height=6cm,
        xmode=linear,
        ymode=log,
        xlabel = {Epoch},
        ylabel = {Loss},
        xmin = 0, xmax = 5200,
        xtick = {0,1000,2000,3000,4000,5000},
        ymin = 1e-6, ymax = 1,
        ytick = {1e0,1e-1,1e-2,1e-3,1e-4,1e-5,1e-6,1e-7},
        legend cell align={left},
        legend style={font=\scriptsize, at={(0.97, 0.97)},anchor=north east},
        %axis line style={draw=none},
        %tick style={draw=none},
        %x tick label style={/pgf/number format/.cd, fixed, fixed zerofill, precision=0, /tikz/.cd},
        %y tick label style={/pgf/number format/.cd, fixed, fixed zerofill, precision=1, /tikz/.cd},
    ]
        
        \addplot[line width=1pt, color=Set1-A] table[x=epoch, y=train_loss_total, col sep=comma]{./figs/data/BBI0.01_beta5_NN0.csv};
        
        \addplot[line width=1pt, color=Set1-B] table[x=epoch, y=bc_loss_train, col sep=comma]{./figs/data/BBI0.01_beta5_NN0.csv};
        \addplot[line width=1pt, color=Set1-C] table[x=epoch, y=bulk_loss_train, col sep=comma]{./figs/data/BBI0.01_beta5_NN0.csv};
        \addplot[line width=1pt, color=Set1-D] table[x=epoch, y=bcp_loss_train, col sep=comma]{./figs/data/BBI0.01_beta5_NN0.csv};

        \addplot[line width=1pt, color=Set1-A, densely dotted] table[x=epoch, y=test_loss_total, col sep=comma]{./figs/data/BBI0.01_beta5_NN0.csv};
        
        \addplot[line width=1pt, color=Set1-B, densely dotted] table[x=epoch, y=bc_loss_test, col sep=comma]{./figs/data/BBI0.01_beta5_NN0.csv};
        
        \addplot[line width=1pt, color=Set1-C, densely dotted] table[x=epoch, y=bulk_loss_test, col sep=comma]{./figs/data/BBI0.01_beta5_NN0.csv};
        
        \addplot[line width=1pt, color=Set1-D, densely dotted] table[x=epoch, y=bcp_loss_test, col sep=comma]{./figs/data/BBI0.01_beta5_NN0.csv};

    \end{axis}
\end{tikzpicture}}}
    \subfloat[$\beta=5$, Network 1]{\adjustbox{width=0.42\linewidth,valign=b}{\begin{tikzpicture}
    \begin{axis}
    [
        axis line style={latex-latex},
        axis y line=left,
        axis x line=left,
        width=9cm,
        height=6cm,
        xmode=linear,
        ymode=log,
        xlabel = {Epoch},
        ylabel = {Loss},
        xmin = 0, xmax = 5200,
        xtick = {0,1000,2000,3000,4000,5000},
        ymin = 1e-6, ymax = 1,
        ytick = {1e0,1e-1,1e-2,1e-3,1e-4,1e-5,1e-6,1e-7},
        legend cell align={left},
        legend style={font=\scriptsize, at={(0.97, 0.97)},anchor=north east},
        %axis line style={draw=none},
        %tick style={draw=none},
        %x tick label style={/pgf/number format/.cd, fixed, fixed zerofill, precision=0, /tikz/.cd},
        %y tick label style={/pgf/number format/.cd, fixed, fixed zerofill, precision=1, /tikz/.cd},
    ]
        
        \addplot[line width=1pt, color=Set1-A] table[x=epoch, y=train_loss_total, col sep=comma]{./figs/data/BBI0.01_beta5_NN1.csv};
        
        \addplot[line width=1pt, color=Set1-B] table[x=epoch, y=bc_loss_train, col sep=comma]{./figs/data/BBI0.01_beta5_NN1.csv};
        \addplot[line width=1pt, color=Set1-C] table[x=epoch, y=bulk_loss_train, col sep=comma]{./figs/data/BBI0.01_beta5_NN1.csv};
        \addplot[line width=1pt, color=Set1-D] table[x=epoch, y=bcp_loss_train, col sep=comma]{./figs/data/BBI0.01_beta5_NN1.csv};

        \addplot[line width=1pt, color=Set1-A, densely dotted] table[x=epoch, y=test_loss_total, col sep=comma]{./figs/data/BBI0.01_beta5_NN1.csv};
        
        \addplot[line width=1pt, color=Set1-B, densely dotted] table[x=epoch, y=bc_loss_test, col sep=comma]{./figs/data/BBI0.01_beta5_NN1.csv};
        \addplot[line width=1pt, color=Set1-C,  densely dotted] table[x=epoch, y=bulk_loss_test, col sep=comma]{./figs/data/BBI0.01_beta5_NN1.csv};
        \addplot[line width=1pt, color=Set1-D, densely dotted] table[x=epoch, y=bcp_loss_test, col sep=comma]{./figs/data/BBI0.01_beta5_NN1.csv};

    \end{axis}
\end{tikzpicture}}}
    \\
    \subfloat[$\beta=15$, Network 2]{\adjustbox{width=0.42\linewidth,valign=b}{\begin{tikzpicture}
    \begin{axis}
    [
        axis line style={latex-latex},
        axis y line=left,
        axis x line=left,
        width=9cm,
        height=6cm,
        xmode=linear,
        ymode=log,
        xlabel = {Epoch},
        ylabel = {Loss},
        xmin = 0, xmax = 5200,
        xtick = {0,1000,2000,3000,4000,5000},
        ymin = 1e-6, ymax = 1,
        ytick = {1e0,1e-1,1e-2,1e-3,1e-4,1e-5,1e-6,1e-7},
        legend cell align={left},
        legend style={font=\scriptsize, at={(0.97, 0.97)},anchor=north east},
        %axis line style={draw=none},
        %tick style={draw=none},
        %x tick label style={/pgf/number format/.cd, fixed, fixed zerofill, precision=0, /tikz/.cd},
        %y tick label style={/pgf/number format/.cd, fixed, fixed zerofill, precision=1, /tikz/.cd},
    ]
        
        \addplot[line width=1pt, color=Set1-A] table[x=epoch, y=train_loss_total, col sep=comma]{./figs/data/BBI0.01_beta15_NN0.csv};
        
        \addplot[line width=1pt, color=Set1-B] table[x=epoch, y=bc_loss_train, col sep=comma]{./figs/data/BBI0.01_beta15_NN0.csv};
        \addplot[line width=1pt, color=Set1-C] table[x=epoch, y=bulk_loss_train, col sep=comma]{./figs/data/BBI0.01_beta15_NN0.csv};
        \addplot[line width=1pt, color=Set1-D] table[x=epoch, y=bcp_loss_train, col sep=comma]{./figs/data/BBI0.01_beta15_NN0.csv};

        \addplot[line width=1pt, color=Set1-A, densely dotted] table[x=epoch, y=test_loss_total, col sep=comma]{./figs/data/BBI0.01_beta15_NN0.csv};
        
        \addplot[line width=1pt, color=Set1-B, densely dotted] table[x=epoch, y=bc_loss_test, col sep=comma]{./figs/data/BBI0.01_beta15_NN0.csv};
        \addplot[line width=1pt, color=Set1-C, densely dotted] table[x=epoch, y=bulk_loss_test, col sep=comma]{./figs/data/BBI0.01_beta15_NN0.csv};
        \addplot[line width=1pt, color=Set1-D, densely dotted] table[x=epoch, y=bcp_loss_test, col sep=comma]{./figs/data/BBI0.01_beta15_NN0.csv};

    \end{axis}
\end{tikzpicture}}}
    \subfloat[$\beta=15$, Network 2]{\adjustbox{width=0.42\linewidth,valign=b}{\begin{tikzpicture}
    \begin{axis}
    [
        axis line style={latex-latex},
        axis y line=left,
        axis x line=left,
        width=9cm,
        height=6cm,
        xmode=linear,
        ymode=log,
        xlabel = {Epoch},
        ylabel = {Loss},
        xmin = 0, xmax = 5200,
        xtick = {0,1000,2000,3000,4000,5000},
        ymin = 1e-6, ymax = 1,
        ytick = {1e0,1e-1,1e-2,1e-3,1e-4,1e-5,1e-6,1e-7},
        legend cell align={left},
        legend style={font=\scriptsize, at={(0.97, 0.97)},anchor=north east},
        %axis line style={draw=none},
        %tick style={draw=none},
        %x tick label style={/pgf/number format/.cd, fixed, fixed zerofill, precision=0, /tikz/.cd},
        %y tick label style={/pgf/number format/.cd, fixed, fixed zerofill, precision=1, /tikz/.cd},
    ]
        
        \addplot[line width=1pt, color=Set1-A] table[x=epoch, y=train_loss_total, col sep=comma]{./figs/data/BBI0.01_beta15_NN1.csv};
        
        \addplot[line width=1pt, color=Set1-B] table[x=epoch, y=bc_loss_train, col sep=comma]{./figs/data/BBI0.01_beta15_NN1.csv};
        \addplot[line width=1pt, color=Set1-C] table[x=epoch, y=bulk_loss_train, col sep=comma]{./figs/data/BBI0.01_beta15_NN1.csv};
        \addplot[line width=1pt, color=Set1-D] table[x=epoch, y=bcp_loss_train, col sep=comma]{./figs/data/BBI0.01_beta15_NN1.csv};

        \addplot[line width=1pt, color=Set1-A, densely dotted] table[x=epoch, y=test_loss_total, col sep=comma]{./figs/data/BBI0.01_beta15_NN1.csv};
        
        \addplot[line width=1pt, color=Set1-B, densely dotted] table[x=epoch, y=bc_loss_test, col sep=comma]{./figs/data/BBI0.01_beta15_NN1.csv};
        \addplot[line width=1pt, color=Set1-C, densely dotted] table[x=epoch, y=bulk_loss_test, col sep=comma]{./figs/data/BBI0.01_beta15_NN1.csv};
        \addplot[line width=1pt, color=Set1-D, densely dotted] table[x=epoch, y=bcp_loss_test, col sep=comma]{./figs/data/BBI0.01_beta15_NN1.csv};

    \end{axis}
\end{tikzpicture}}}
    \\
    \subfloat[$\beta=30$, Network 2]{\adjustbox{width=0.42\linewidth,valign=b}{\begin{tikzpicture}
    \begin{axis}
    [
        axis line style={latex-latex},
        axis y line=left,
        axis x line=left,
        width=9cm,
        height=6cm,
        xmode=linear,
        ymode=log,
        xlabel = {Epoch},
        ylabel = {Loss},
        xmin = 0, xmax = 5200,
        xtick = {0,1000,2000,3000,4000,5000},
        ymin = 1e-6, ymax = 1,
        ytick = {1e0,1e-1,1e-2,1e-3,1e-4,1e-5,1e-6,1e-7},
        legend cell align={left},
        legend style={font=\scriptsize, at={(0.97, 0.97)},anchor=north east},
        %axis line style={draw=none},
        %tick style={draw=none},
        %x tick label style={/pgf/number format/.cd, fixed, fixed zerofill, precision=0, /tikz/.cd},
        %y tick label style={/pgf/number format/.cd, fixed, fixed zerofill, precision=1, /tikz/.cd},
    ]
        
        \addplot[line width=1pt, color=Set1-A] table[x=epoch, y=train_loss_total, col sep=comma]{./figs/data/BBI0.01_beta30_NN0.csv};
        
        \addplot[line width=1pt, color=Set1-B] table[x=epoch, y=bc_loss_train, col sep=comma]{./figs/data/BBI0.01_beta30_NN0.csv};
        \addplot[line width=1pt, color=Set1-C] table[x=epoch, y=bulk_loss_train, col sep=comma]{./figs/data/BBI0.01_beta30_NN0.csv};
        \addplot[line width=1pt, color=Set1-D] table[x=epoch, y=bcp_loss_train, col sep=comma]{./figs/data/BBI0.01_beta30_NN0.csv};

        \addplot[line width=1pt, color=Set1-A, densely dotted] table[x=epoch, y=test_loss_total, col sep=comma]{./figs/data/BBI0.01_beta30_NN0.csv};
        
        \addplot[line width=1pt, color=Set1-B, densely dotted] table[x=epoch, y=bc_loss_test, col sep=comma]{./figs/data/BBI0.01_beta30_NN0.csv};
        \addplot[line width=1pt, color=Set1-C, densely dotted] table[x=epoch, y=bulk_loss_test, col sep=comma]{./figs/data/BBI0.01_beta30_NN0.csv};
        \addplot[line width=1pt, color=Set1-D, densely dotted] table[x=epoch, y=bcp_loss_test, col sep=comma]{./figs/data/BBI0.01_beta30_NN0.csv};

    \end{axis}
\end{tikzpicture}}}
    \subfloat[$\beta=30$, Network 2]{\adjustbox{width=0.42\linewidth,valign=b}{\begin{tikzpicture}
    \begin{axis}
    [
        axis line style={latex-latex},
        axis y line=left,
        axis x line=left,
        width=9cm,
        height=6cm,
        xmode=linear,
        ymode=log,
        xlabel = {Epoch},
        ylabel = {Loss},
        xmin = 0, xmax = 5200,
        xtick = {0,1000,2000,3000,4000,5000},
        ymin = 1e-6, ymax = 1,
        ytick = {1e0,1e-1,1e-2,1e-3,1e-4,1e-5,1e-6,1e-7},
        legend cell align={left},
        legend style={font=\scriptsize, at={(0.97, 0.97)},anchor=north east},
        %axis line style={draw=none},
        %tick style={draw=none},
        %x tick label style={/pgf/number format/.cd, fixed, fixed zerofill, precision=0, /tikz/.cd},
        %y tick label style={/pgf/number format/.cd, fixed, fixed zerofill, precision=1, /tikz/.cd},
    ]
        
        \addplot[line width=1pt, color=Set1-A] table[x=epoch, y=train_loss_total, col sep=comma]{./figs/data/BBI0.01_beta30_NN1.csv};
        
        \addplot[line width=1pt, color=Set1-B] table[x=epoch, y=bc_loss_train, col sep=comma]{./figs/data/BBI0.01_beta30_NN1.csv};
        
        \addplot[line width=1pt, color=Set1-C] table[x=epoch, y=bulk_loss_train, col sep=comma]{./figs/data/BBI0.01_beta30_NN1.csv};
        \addplot[line width=1pt, color=Set1-D] table[x=epoch, y=bcp_loss_train, col sep=comma]{./figs/data/BBI0.01_beta30_NN1.csv};

        \addplot[line width=1pt, color=Set1-A] table[x=epoch, y=test_loss_total, col sep=comma]{./figs/data/BBI0.01_beta30_NN1.csv};
        
        \addplot[line width=1pt, color=Set1-B, densely dotted] table[x=epoch, y=bc_loss_test, col sep=comma]{./figs/data/BBI0.01_beta30_NN1.csv};
        
        \addplot[line width=1pt, color=Set1-C, densely dotted] table[x=epoch, y=bulk_loss_test, col sep=comma]{./figs/data/BBI0.01_beta30_NN1.csv};
        
        \addplot[line width=1pt, color=Set1-D, densely dotted] table[x=epoch, y=bcp_loss_test, col sep=comma]{./figs/data/BBI0.01_beta30_NN1.csv};

    \end{axis}
\end{tikzpicture}}}
    \caption{\label{fig:bbi_loss} Median loss versus epoch over 10 samples when using BBI. \emph{Solid} line is the total, and the other lines show the separate contributions. \emph{dashed} for initial condition, \emph{dotted} for the bulk, and \emph{dash-dotted} for the periodic boundary.}
\end{figure}

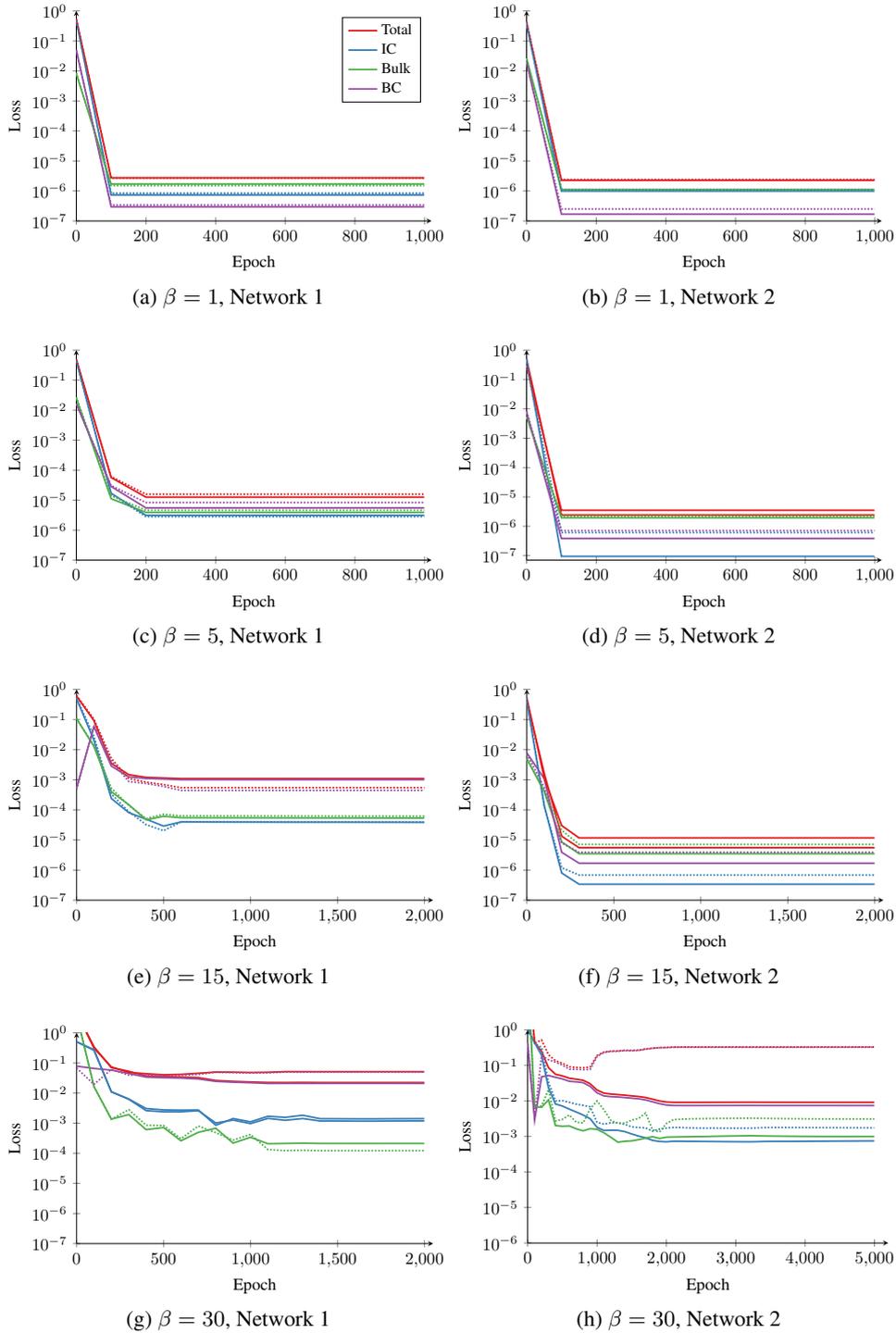
\begin{figure}[tbhp]
    \centering
    \subfloat[$\beta=1$, Network 1]{\adjustbox{width=0.42\linewidth,valign=b}{\begin{tikzpicture}
    \begin{axis}
    [
        axis line style={latex-latex},
        axis y line=left,
        axis x line=left,
        width=9cm,
        height=6cm,
        xmode=linear,
        ymode=log,
        xlabel = {Epoch},
        ylabel = {Loss},
        xmin = 0, xmax = 1020,
        xtick = {0,200,400,600,800,1000},
        ymin = 1e-7, ymax = 1,
        ytick = {1e0,1e-1,1e-2,1e-3,1e-4,1e-5,1e-6,1e-7},
        legend cell align={left},
        legend style={font=\small, at={(0.97, 0.97)},anchor=north east},
        %axis line style={draw=none},
        %tick style={draw=none},
        %x tick label style={/pgf/number format/.cd, fixed, fixed zerofill, precision=0, /tikz/.cd},
        %y tick label style={/pgf/number format/.cd, fixed, fixed zerofill, precision=1, /tikz/.cd},
    ]
        
        \addplot[line width=1pt, color=Set1-A] table[x=epoch, y=train_loss_total, col sep=comma]{./figs/data/LBFGS0.1_beta1_NN0.csv};
        \addplot[line width=1pt, color=Set1-B] table[x=epoch, y=bc_loss_train, col sep=comma]{./figs/data/LBFGS0.1_beta1_NN0.csv};
        \addplot[line width=1pt, color=Set1-C] table[x=epoch, y=bulk_loss_train, col sep=comma]{./figs/data/LBFGS0.1_beta1_NN0.csv};
        \addplot[line width=1pt, color=Set1-D] table[x=epoch, y=bcp_loss_train, col sep=comma]{./figs/data/LBFGS0.1_beta1_NN0.csv};

        \addplot[line width=1pt, color=Set1-A, densely dotted] table[x=epoch, y=test_loss_total, col sep=comma]{./figs/data/LBFGS0.1_beta1_NN0.csv};
        \addplot[line width=1pt, color=Set1-B, densely dotted] table[x=epoch, y=bc_loss_test, col sep=comma]{./figs/data/LBFGS0.1_beta1_NN0.csv};
        \addplot[line width=1pt, color=Set1-C, densely dotted] table[x=epoch, y=bulk_loss_test, col sep=comma]{./figs/data/LBFGS0.1_beta1_NN0.csv};
        \addplot[line width=1pt, color=Set1-D, densely dotted] table[x=epoch, y=bcp_loss_test, col sep=comma]{./figs/data/LBFGS0.1_beta1_NN0.csv};

        \addlegendentry{Total};
        \addlegendentry{IC};
        \addlegendentry{Bulk};
        \addlegendentry{BC};

    \end{axis}
\end{tikzpicture}}}
    \subfloat[$\beta=1$, Network 2]{\adjustbox{width=0.42\linewidth,valign=b}{\begin{tikzpicture}
    \begin{axis}
    [
        axis line style={latex-latex},
        axis y line=left,
        axis x line=left,
        width=9cm,
        height=6cm,
        xmode=linear,
        ymode=log,
        xlabel = {Epoch},
        ylabel = {Loss},
        xmin = 0, xmax = 1020,
        xtick = {0,200,400,600,800,1000},
        ymin = 1e-7, ymax = 1,
        ytick = {1e0,1e-1,1e-2,1e-3,1e-4,1e-5,1e-6,1e-7},
        legend cell align={left},
        legend style={font=\scriptsize, at={(0.97, 0.97)},anchor=north east},
        %axis line style={draw=none},
        %tick style={draw=none},
        %x tick label style={/pgf/number format/.cd, fixed, fixed zerofill, precision=0, /tikz/.cd},
        %y tick label style={/pgf/number format/.cd, fixed, fixed zerofill, precision=1, /tikz/.cd},
    ]
        
        \addplot[line width=1pt, color=Set1-A] table[x=epoch, y=train_loss_total, col sep=comma]{./figs/data/LBFGS0.1_beta1_NN1.csv};
        
        \addplot[line width=1pt, color=Set1-B] table[x=epoch, y=bc_loss_train, col sep=comma]{./figs/data/LBFGS0.1_beta1_NN1.csv};
        \addplot[line width=1pt, color=Set1-C] table[x=epoch, y=bulk_loss_train, col sep=comma]{./figs/data/LBFGS0.1_beta1_NN1.csv};
        \addplot[line width=1pt, color=Set1-D] table[x=epoch, y=bcp_loss_train, col sep=comma]{./figs/data/LBFGS0.1_beta1_NN1.csv};

        \addplot[line width=1pt, color=Set1-A, densely dotted] table[x=epoch, y=test_loss_total, col sep=comma]{./figs/data/LBFGS0.1_beta1_NN1.csv};
        
        \addplot[line width=1pt, color=Set1-B, densely dotted] table[x=epoch, y=bc_loss_test, col sep=comma]{./figs/data/LBFGS0.1_beta1_NN1.csv};
        \addplot[line width=1pt, color=Set1-C, densely dotted] table[x=epoch, y=bulk_loss_test, col sep=comma]{./figs/data/LBFGS0.1_beta1_NN1.csv};
        \addplot[line width=1pt, color=Set1-D, densely dotted] table[x=epoch, y=bcp_loss_test, col sep=comma]{./figs/data/LBFGS0.1_beta1_NN1.csv};

    \end{axis}
\end{tikzpicture}}}
    \\
    \subfloat[$\beta=5$, Network 1]{\adjustbox{width=0.42\linewidth,valign=b}{\begin{tikzpicture}
    \begin{axis}
    [
        axis line style={latex-latex},
        axis y line=left,
        axis x line=left,
        width=9cm,
        height=6cm,
        xmode=linear,
        ymode=log,
        xlabel = {Epoch},
        ylabel = {Loss},
        xmin = 0, xmax = 1020,
        xtick = {0,200,400,600,800,1000},
        ymin = 1e-7, ymax = 1,
        ytick = {1e0,1e-1,1e-2,1e-3,1e-4,1e-5,1e-6,1e-7},
        legend cell align={left},
        legend style={font=\scriptsize, at={(0.97, 0.97)},anchor=north east},
        %axis line style={draw=none},
        %tick style={draw=none},
        %x tick label style={/pgf/number format/.cd, fixed, fixed zerofill, precision=0, /tikz/.cd},
        %y tick label style={/pgf/number format/.cd, fixed, fixed zerofill, precision=1, /tikz/.cd},
    ]
        
        \addplot[line width=1pt, color=Set1-A] table[x=epoch, y=train_loss_total, col sep=comma]{./figs/data/LBFGS0.1_beta5_NN0.csv};
        
        \addplot[line width=1pt, color=Set1-B] table[x=epoch, y=bc_loss_train, col sep=comma]{./figs/data/LBFGS0.1_beta5_NN0.csv};
        \addplot[line width=1pt, color=Set1-C] table[x=epoch, y=bulk_loss_train, col sep=comma]{./figs/data/LBFGS0.1_beta5_NN0.csv};
        \addplot[line width=1pt, color=Set1-D] table[x=epoch, y=bcp_loss_train, col sep=comma]{./figs/data/LBFGS0.1_beta5_NN0.csv};

        \addplot[line width=1pt, color=Set1-A, densely dotted] table[x=epoch, y=test_loss_total, col sep=comma]{./figs/data/LBFGS0.1_beta5_NN0.csv};
        
        \addplot[line width=1pt, color=Set1-B, densely dotted] table[x=epoch, y=bc_loss_test, col sep=comma]{./figs/data/LBFGS0.1_beta5_NN0.csv};
        \addplot[line width=1pt, color=Set1-C, densely dotted] table[x=epoch, y=bulk_loss_test, col sep=comma]{./figs/data/LBFGS0.1_beta5_NN0.csv};
        \addplot[line width=1pt, color=Set1-D, densely dotted] table[x=epoch, y=bcp_loss_test, col sep=comma]{./figs/data/LBFGS0.1_beta5_NN0.csv};

    \end{axis}
\end{tikzpicture}}}
    \subfloat[$\beta=5$, Network 2]{\adjustbox{width=0.42\linewidth,valign=b}{\begin{tikzpicture}
    \begin{axis}
    [
        axis line style={latex-latex},
        axis y line=left,
        axis x line=left,
        width=9cm,
        height=6cm,
        xmode=linear,
        ymode=log,
        xlabel = {Epoch},
        ylabel = {Loss},
        xmin = 0, xmax = 1020,
        xtick = {0,200,400,600,800,1000},
        ymin = 7e-8, ymax = 1,
        ytick = {1e0,1e-1,1e-2,1e-3,1e-4,1e-5,1e-6,1e-7},
        legend cell align={left},
        legend style={font=\scriptsize, at={(0.97, 0.97)},anchor=north east},
        %axis line style={draw=none},
        %tick style={draw=none},
        %x tick label style={/pgf/number format/.cd, fixed, fixed zerofill, precision=0, /tikz/.cd},
        %y tick label style={/pgf/number format/.cd, fixed, fixed zerofill, precision=1, /tikz/.cd},
    ]
        
        \addplot[line width=1pt, color=Set1-A] table[x=epoch, y=train_loss_total, col sep=comma]{./figs/data/LBFGS0.1_beta5_NN1.csv};
        
        \addplot[line width=1pt, color=Set1-B] table[x=epoch, y=bc_loss_train, col sep=comma]{./figs/data/LBFGS0.1_beta5_NN1.csv};
        \addplot[line width=1pt, color=Set1-C] table[x=epoch, y=bulk_loss_train, col sep=comma]{./figs/data/LBFGS0.1_beta5_NN1.csv};
        \addplot[line width=1pt, color=Set1-D] table[x=epoch, y=bcp_loss_train, col sep=comma]{./figs/data/LBFGS0.1_beta5_NN1.csv};

        \addplot[line width=1pt, color=Set1-A] table[x=epoch, y=test_loss_total, col sep=comma]{./figs/data/LBFGS0.1_beta5_NN1.csv};
        
        \addplot[line width=1pt, color=Set1-B, densely dotted] table[x=epoch, y=bc_loss_test, col sep=comma]{./figs/data/LBFGS0.1_beta5_NN1.csv};
        \addplot[line width=1pt, color=Set1-C, densely dotted] table[x=epoch, y=bulk_loss_test, col sep=comma]{./figs/data/LBFGS0.1_beta5_NN1.csv};
        \addplot[line width=1pt, color=Set1-D, densely dotted] table[x=epoch, y=bcp_loss_test, col sep=comma]{./figs/data/LBFGS0.1_beta5_NN1.csv};

    \end{axis}
\end{tikzpicture}}}
    \\
    \subfloat[$\beta=15$, Network 1]{\adjustbox{width=0.42\linewidth,valign=b}{\begin{tikzpicture}
    \begin{axis}
    [
        axis line style={latex-latex},
        axis y line=left,
        axis x line=left,
        width=9cm,
        height=6cm,
        xmode=linear,
        ymode=log,
        xlabel = {Epoch},
        ylabel = {Loss},
        xmin = 0, xmax = 2050,
        xtick = {0,500,1000,1500,2000},
        ymin = 1e-7, ymax = 1,
        ytick = {1e0,1e-1,1e-2,1e-3,1e-4,1e-5,1e-6,1e-7},
        legend cell align={left},
        legend style={font=\scriptsize, at={(0.97, 0.97)},anchor=north east},
        %axis line style={draw=none},
        %tick style={draw=none},
        %x tick label style={/pgf/number format/.cd, fixed, fixed zerofill, precision=0, /tikz/.cd},
        %y tick label style={/pgf/number format/.cd, fixed, fixed zerofill, precision=1, /tikz/.cd},
    ]
        
        \addplot[line width=1pt, color=Set1-A] table[x=epoch, y=train_loss_total, col sep=comma]{./figs/data/LBFGS0.01_beta15_NN0.csv};
        
        \addplot[line width=1pt, color=Set1-B] table[x=epoch, y=bc_loss_train, col sep=comma]{./figs/data/LBFGS0.01_beta15_NN0.csv};
        \addplot[line width=1pt, color=Set1-C] table[x=epoch, y=bulk_loss_train, col sep=comma]{./figs/data/LBFGS0.01_beta15_NN0.csv};
        \addplot[line width=1pt, color=Set1-D] table[x=epoch, y=bcp_loss_train, col sep=comma]{./figs/data/LBFGS0.01_beta15_NN0.csv};

        \addplot[line width=1pt, color=Set1-A,  densely dotted] table[x=epoch, y=test_loss_total, col sep=comma]{./figs/data/LBFGS0.01_beta15_NN0.csv};
        
        \addplot[line width=1pt, color=Set1-B, densely dotted] table[x=epoch, y=bc_loss_test, col sep=comma]{./figs/data/LBFGS0.01_beta15_NN0.csv};
        \addplot[line width=1pt, color=Set1-C, densely dotted] table[x=epoch, y=bulk_loss_test, col sep=comma]{./figs/data/LBFGS0.01_beta15_NN0.csv};
        \addplot[line width=1pt, color=Set1-D, densely dotted] table[x=epoch, y=bcp_loss_test, col sep=comma]{./figs/data/LBFGS0.01_beta15_NN0.csv};

    \end{axis}
\end{tikzpicture}}}
    \subfloat[$\beta=15$, Network 2]{\adjustbox{width=0.42\linewidth,valign=b}{\begin{tikzpicture}
    \begin{axis}
    [
        axis line style={latex-latex},
        axis y line=left,
        axis x line=left,
        width=9cm,
        height=6cm,
        xmode=linear,
        ymode=log,
        xlabel = {Epoch},
        ylabel = {Loss},
        xmin = 0, xmax = 2050,
        xtick = {0,500,1000,1500,2000},
        ymin = 1e-7, ymax = 1,
        ytick = {1e0,1e-1,1e-2,1e-3,1e-4,1e-5,1e-6,1e-7},
        legend cell align={left},
        legend style={font=\scriptsize, at={(0.97, 0.97)},anchor=north east},
        %axis line style={draw=none},
        %tick style={draw=none},
        %x tick label style={/pgf/number format/.cd, fixed, fixed zerofill, precision=0, /tikz/.cd},
        %y tick label style={/pgf/number format/.cd, fixed, fixed zerofill, precision=1, /tikz/.cd},
    ]
        
        \addplot[line width=1pt, color=Set1-A] table[x=epoch, y=train_loss_total, col sep=comma]{./figs/data/LBFGS1_beta15_NN1.csv};
        
        \addplot[line width=1pt, color=Set1-B] table[x=epoch, y=bc_loss_train, col sep=comma]{./figs/data/LBFGS1_beta15_NN1.csv};
        \addplot[line width=1pt, color=Set1-C] table[x=epoch, y=bulk_loss_train, col sep=comma]{./figs/data/LBFGS1_beta15_NN1.csv};
        \addplot[line width=1pt, color=Set1-D] table[x=epoch, y=bcp_loss_train, col sep=comma]{./figs/data/LBFGS1_beta15_NN1.csv};

        \addplot[line width=1pt, color=Set1-A] table[x=epoch, y=test_loss_total, col sep=comma]{./figs/data/LBFGS1_beta15_NN1.csv};
        
        \addplot[line width=1pt, color=Set1-B, densely dotted] table[x=epoch, y=bc_loss_test, col sep=comma]{./figs/data/LBFGS1_beta15_NN1.csv};
        \addplot[line width=1pt, color=Set1-C, densely dotted] table[x=epoch, y=bulk_loss_test, col sep=comma]{./figs/data/LBFGS1_beta15_NN1.csv};
        \addplot[line width=1pt, color=Set1-D, densely dotted] table[x=epoch, y=bcp_loss_test, col sep=comma]{./figs/data/LBFGS1_beta15_NN1.csv};

    \end{axis}
\end{tikzpicture}}}
    \\
    \subfloat[$\beta=30$, Network 1]{\adjustbox{width=0.42\linewidth,valign=b}{\begin{tikzpicture}
    \begin{axis}
    [
        axis line style={latex-latex},
        axis y line=left,
        axis x line=left,
        width=9cm,
        height=6cm,
        xmode=linear,
        ymode=log,
        xlabel = {Epoch},
        ylabel = {Loss},
        xmin = 0, xmax = 2050,
        xtick = {0,500,1000,1500,2000},
        ymin = 1e-7, ymax = 1,
        ytick = {1e0,1e-1,1e-2,1e-3,1e-4,1e-5,1e-6,1e-7},
        legend cell align={left},
        legend style={font=\scriptsize, at={(0.97, 0.97)},anchor=north east},
        %axis line style={draw=none},
        %tick style={draw=none},
        %x tick label style={/pgf/number format/.cd, fixed, fixed zerofill, precision=0, /tikz/.cd},
        %y tick label style={/pgf/number format/.cd, fixed, fixed zerofill, precision=1, /tikz/.cd},
    ]
        
        \addplot[line width=1pt, color=Set1-A] table[x=epoch, y=train_loss_total, col sep=comma]{./figs/data/LBFGS0.01_beta30_NN0.csv};
        
        \addplot[line width=1pt, color=Set1-B] table[x=epoch, y=bc_loss_train, col sep=comma]{./figs/data/LBFGS0.01_beta30_NN0.csv};
        \addplot[line width=1pt, color=Set1-C] table[x=epoch, y=bulk_loss_train, col sep=comma]{./figs/data/LBFGS0.01_beta30_NN0.csv};
        \addplot[line width=1pt, color=Set1-D] table[x=epoch, y=bcp_loss_train, col sep=comma]{./figs/data/LBFGS0.01_beta30_NN0.csv};

        \addplot[line width=1pt, color=Set1-A] table[x=epoch, y=test_loss_total, col sep=comma]{./figs/data/LBFGS0.01_beta30_NN0.csv};
        
        \addplot[line width=1pt, color=Set1-B] table[x=epoch, y=bc_loss_test, col sep=comma]{./figs/data/LBFGS0.01_beta30_NN0.csv};
        \addplot[line width=1pt, color=Set1-C, densely dotted] table[x=epoch, y=bulk_loss_test, col sep=comma]{./figs/data/LBFGS0.01_beta30_NN0.csv};
        \addplot[line width=1pt, color=Set1-D, densely dotted] table[x=epoch, y=bcp_loss_test, col sep=comma]{./figs/data/LBFGS0.01_beta30_NN0.csv};

    \end{axis}
\end{tikzpicture}}}
    \subfloat[$\beta=30$, Network 2]{\adjustbox{width=0.42\linewidth,valign=b}{\begin{tikzpicture}
    \begin{axis}
    [
        axis line style={latex-latex},
        axis y line=left,
        axis x line=left,
        width=9cm,
        height=6cm,
        xmode=linear,
        ymode=log,
        xlabel = {Epoch},
        ylabel = {Loss},
        xmin = 0, xmax = 5200,
        xtick = {0,1000,2000,3000,4000,5000},
        ymin = 1e-6, ymax = 1,
        ytick = {1e0,1e-1,1e-2,1e-3,1e-4,1e-5,1e-6,1e-7},
        legend cell align={left},
        legend style={font=\scriptsize, at={(0.97, 0.97)},anchor=north east},
        %axis line style={draw=none},
        %tick style={draw=none},
        %x tick label style={/pgf/number format/.cd, fixed, fixed zerofill, precision=0, /tikz/.cd},
        %y tick label style={/pgf/number format/.cd, fixed, fixed zerofill, precision=1, /tikz/.cd},
    ]
        
        \addplot[line width=1pt, color=Set1-A] table[x=epoch, y=train_loss_total, col sep=comma]{./figs/data/LBFGS0.001_beta30_NN1.csv};
        
        \addplot[line width=1pt, color=Set1-B] table[x=epoch, y=bc_loss_train, col sep=comma]{./figs/data/LBFGS0.001_beta30_NN1.csv};
        \addplot[line width=1pt, color=Set1-C] table[x=epoch, y=bulk_loss_train, col sep=comma]{./figs/data/LBFGS0.001_beta30_NN1.csv};
        \addplot[line width=1pt, color=Set1-D] table[x=epoch, y=bcp_loss_train, col sep=comma]{./figs/data/LBFGS0.001_beta30_NN1.csv};

        \addplot[line width=1pt, color=Set1-A, densely dotted] table[x=epoch, y=test_loss_total, col sep=comma]{./figs/data/LBFGS0.001_beta30_NN1.csv};
        
        \addplot[line width=1pt, color=Set1-B, densely dotted] table[x=epoch, y=bc_loss_test, col sep=comma]{./figs/data/LBFGS0.001_beta30_NN1.csv};
        \addplot[line width=1pt, color=Set1-C, densely dotted] table[x=epoch, y=bulk_loss_test, col sep=comma]{./figs/data/LBFGS0.001_beta30_NN1.csv};
        \addplot[line width=1pt, color=Set1-D, densely dotted] table[x=epoch, y=bcp_loss_test, col sep=comma]{./figs/data/LBFGS0.001_beta30_NN1.csv};

    \end{axis}
\end{tikzpicture}}}
    \caption{\label{fig:lbfgd_loss} Median loss versus epoch over 10 samples when using LBFGS. \emph{Solid} line is the total, and the other lines show the separate contributions. \emph{dashed} for initial condition, \emph{dotted} for the bulk, and \emph{dash-dotted} for the periodic boundary. Note the changing number of epochs for convergence.}
\end{figure}

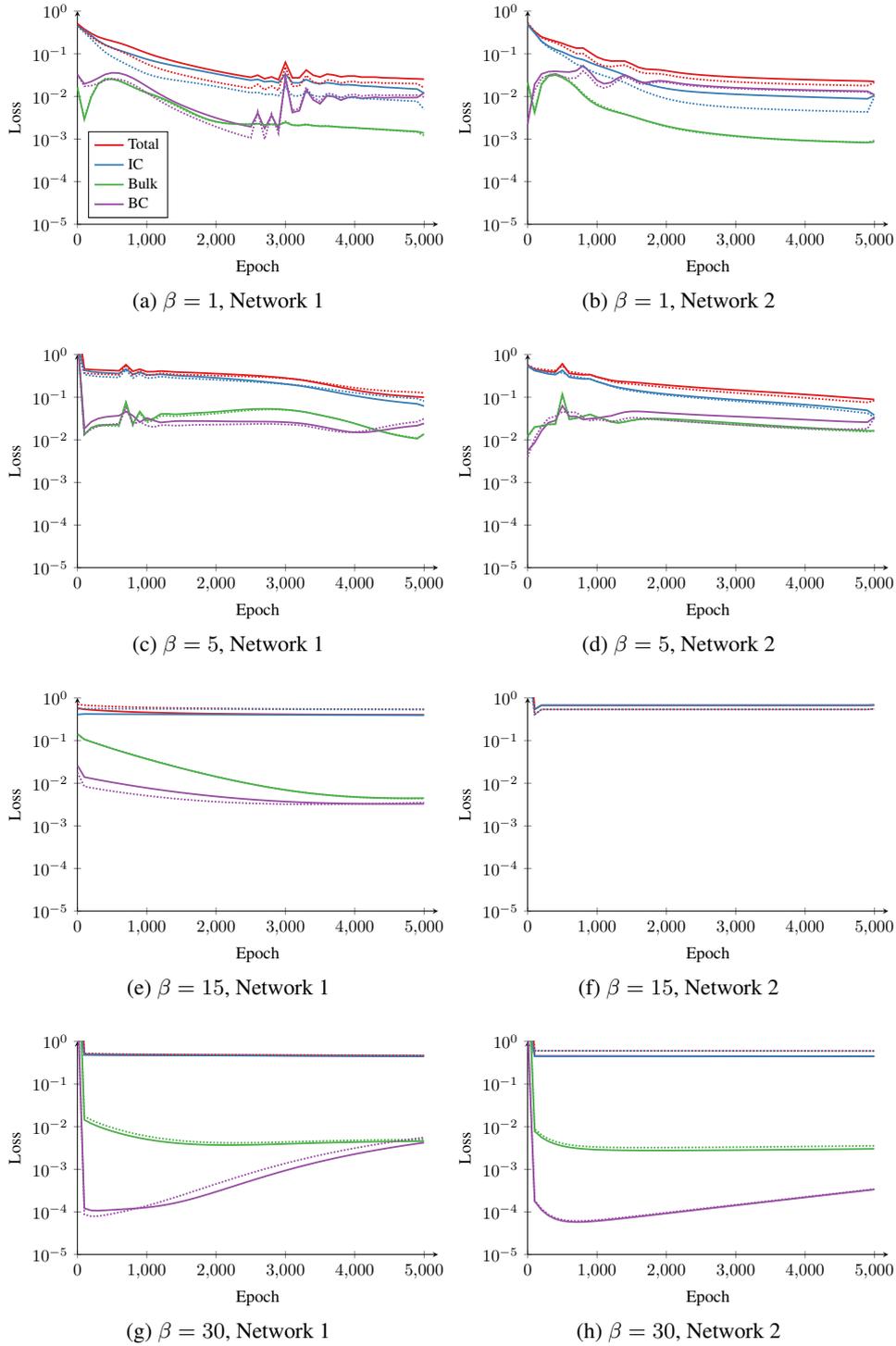
\begin{figure}[tbhp]
    \centering
    \subfloat[$\beta=1$, Network 1]{\adjustbox{width=0.42\linewidth,valign=b}{\begin{tikzpicture}
    \begin{axis}
    [
        axis line style={latex-latex},
        axis y line=left,
        axis x line=left,
        width=9cm,
        height=6cm,
        xmode=linear,
        ymode=log,
        xlabel = {Epoch},
        ylabel = {Loss},
        xmin = 0, xmax = 5200,
        xtick = {0,1000,2000,3000,4000,5000},
        ymin = 1e-5, ymax = 1,
        ytick = {1e0,1e-1,1e-2,1e-3,1e-4,1e-5,1e-6,1e-7},
        legend cell align={left},
        legend style={font=\small, at={(0.03, 0.03)},anchor=south west},
        %axis line style={draw=none},
        %tick style={draw=none},
        %x tick label style={/pgf/number format/.cd, fixed, fixed zerofill, precision=0, /tikz/.cd},
        %y tick label style={/pgf/number format/.cd, fixed, fixed zerofill, precision=1, /tikz/.cd},
    ]
        
        \addplot[line width=1pt, color=Set1-A] table[x=epoch, y=train_loss_total, col sep=comma]{./figs/data/SGD0.01_beta1_NN0.csv};
        \addplot[line width=1pt, color=Set1-B] table[x=epoch, y=bc_loss_train, col sep=comma]{./figs/data/SGD0.01_beta1_NN0.csv};
        \addplot[line width=1pt, color=Set1-C] table[x=epoch, y=bulk_loss_train, col sep=comma]{./figs/data/SGD0.01_beta1_NN0.csv};
        \addplot[line width=1pt, color=Set1-D] table[x=epoch, y=bcp_loss_train, col sep=comma]{./figs/data/SGD0.01_beta1_NN0.csv};

        \addplot[line width=1pt, color=Set1-A, densely dotted] table[x=epoch, y=test_loss_total, col sep=comma]{./figs/data/SGD0.01_beta1_NN0.csv};
        \addplot[line width=1pt, color=Set1-B, densely dotted] table[x=epoch, y=bc_loss_test, col sep=comma]{./figs/data/SGD0.01_beta1_NN0.csv};
        \addplot[line width=1pt, color=Set1-C, densely dotted] table[x=epoch, y=bulk_loss_test, col sep=comma]{./figs/data/SGD0.01_beta1_NN0.csv};
        \addplot[line width=1pt, color=Set1-D, densely dotted] table[x=epoch, y=bcp_loss_test, col sep=comma]{./figs/data/SGD0.01_beta1_NN0.csv};
        
        \addlegendentry{Total};
        \addlegendentry{IC};
        \addlegendentry{Bulk};
        \addlegendentry{BC};

    \end{axis}
\end{tikzpicture}}}
    \subfloat[$\beta=1$, Network 2]{\adjustbox{width=0.42\linewidth,valign=b}{\begin{tikzpicture}
    \begin{axis}
    [
        axis line style={latex-latex},
        axis y line=left,
        axis x line=left,
        width=9cm,
        height=6cm,
        xmode=linear,
        ymode=log,
        xlabel = {Epoch},
        ylabel = {Loss},
        xmin = 0, xmax = 5200,
        xtick = {0,1000,2000,3000,4000,5000},
        ymin = 1e-5, ymax = 1,
        ytick = {1e0,1e-1,1e-2,1e-3,1e-4,1e-5,1e-6,1e-7},
        legend cell align={left},
        legend style={font=\scriptsize, at={(0.97, 0.97)},anchor=north east},
        %axis line style={draw=none},
        %tick style={draw=none},
        %x tick label style={/pgf/number format/.cd, fixed, fixed zerofill, precision=0, /tikz/.cd},
        %y tick label style={/pgf/number format/.cd, fixed, fixed zerofill, precision=1, /tikz/.cd},
    ]
        
        \addplot[line width=1pt, color=Set1-A] table[x=epoch, y=train_loss_total, col sep=comma]{./figs/data/SGD0.01_beta1_NN1.csv};
        
        \addplot[line width=1pt, color=Set1-B] table[x=epoch, y=bc_loss_train, col sep=comma]{./figs/data/SGD0.01_beta1_NN1.csv};
        \addplot[line width=1pt, color=Set1-C] table[x=epoch, y=bulk_loss_train, col sep=comma]{./figs/data/SGD0.01_beta1_NN1.csv};
        \addplot[line width=1pt, color=Set1-D] table[x=epoch, y=bcp_loss_train, col sep=comma]{./figs/data/SGD0.01_beta1_NN1.csv};

        \addplot[line width=1pt, color=Set1-A, densely dotted] table[x=epoch, y=test_loss_total, col sep=comma]{./figs/data/SGD0.01_beta1_NN1.csv};
        
        \addplot[line width=1pt, color=Set1-B, densely dotted] table[x=epoch, y=bc_loss_test, col sep=comma]{./figs/data/SGD0.01_beta1_NN1.csv};
        \addplot[line width=1pt, color=Set1-C, densely dotted] table[x=epoch, y=bulk_loss_test, col sep=comma]{./figs/data/SGD0.01_beta1_NN1.csv};
        \addplot[line width=1pt, color=Set1-D, densely dotted] table[x=epoch, y=bcp_loss_test, col sep=comma]{./figs/data/SGD0.01_beta1_NN1.csv};

    \end{axis}
\end{tikzpicture}}}
    \\
    \subfloat[$\beta=5$, Network 1]{\adjustbox{width=0.42\linewidth,valign=b}{\begin{tikzpicture}
    \begin{axis}
    [
        axis line style={latex-latex},
        axis y line=left,
        axis x line=left,
        width=9cm,
        height=6cm,
        xmode=linear,
        ymode=log,
        xlabel = {Epoch},
        ylabel = {Loss},
        xmin = 0, xmax = 5200,
        xtick = {0,1000,2000,3000,4000,5000},
        ymin = 1e-5, ymax = 1,
        ytick = {1e0,1e-1,1e-2,1e-3,1e-4,1e-5,1e-6,1e-7},
        legend cell align={left},
        legend style={font=\scriptsize, at={(0.97, 0.97)},anchor=north east},
        %axis line style={draw=none},
        %tick style={draw=none},
        %x tick label style={/pgf/number format/.cd, fixed, fixed zerofill, precision=0, /tikz/.cd},
        %y tick label style={/pgf/number format/.cd, fixed, fixed zerofill, precision=1, /tikz/.cd},
    ]
        
        \addplot[line width=1pt, color=Set1-A] table[x=epoch, y=train_loss_total, col sep=comma]{./figs/data/SGD0.01_beta5_NN0.csv};
        
        \addplot[line width=1pt, color=Set1-B] table[x=epoch, y=bc_loss_train, col sep=comma]{./figs/data/SGD0.01_beta5_NN0.csv};
        \addplot[line width=1pt, color=Set1-C] table[x=epoch, y=bulk_loss_train, col sep=comma]{./figs/data/SGD0.01_beta5_NN0.csv};
        \addplot[line width=1pt, color=Set1-D] table[x=epoch, y=bcp_loss_train, col sep=comma]{./figs/data/SGD0.01_beta5_NN0.csv};

        \addplot[line width=1pt, color=Set1-A, densely dotted] table[x=epoch, y=test_loss_total, col sep=comma]{./figs/data/SGD0.01_beta5_NN0.csv};
        
        \addplot[line width=1pt, color=Set1-B, densely dotted] table[x=epoch, y=bc_loss_test, col sep=comma]{./figs/data/SGD0.01_beta5_NN0.csv};
        \addplot[line width=1pt, color=Set1-C, densely dotted] table[x=epoch, y=bulk_loss_test, col sep=comma]{./figs/data/SGD0.01_beta5_NN0.csv};
        \addplot[line width=1pt, color=Set1-D, densely dotted] table[x=epoch, y=bcp_loss_test, col sep=comma]{./figs/data/SGD0.01_beta5_NN0.csv};

    \end{axis}
\end{tikzpicture}}}
    \subfloat[$\beta=5$, Network 2]{\adjustbox{width=0.42\linewidth,valign=b}{\begin{tikzpicture}
    \begin{axis}
    [
        axis line style={latex-latex},
        axis y line=left,
        axis x line=left,
        width=9cm,
        height=6cm,
        xmode=linear,
        ymode=log,
        xlabel = {Epoch},
        ylabel = {Loss},
        xmin = 0, xmax = 5200,
        xtick = {0,1000,2000,3000,4000,5000},
        ymin = 1e-5, ymax = 1,
        ytick = {1e0,1e-1,1e-2,1e-3,1e-4,1e-5,1e-6,1e-7},
        legend cell align={left},
        legend style={font=\scriptsize, at={(0.97, 0.97)},anchor=north east},
        %axis line style={draw=none},
        %tick style={draw=none},
        %x tick label style={/pgf/number format/.cd, fixed, fixed zerofill, precision=0, /tikz/.cd},
        %y tick label style={/pgf/number format/.cd, fixed, fixed zerofill, precision=1, /tikz/.cd},
    ]
        
        \addplot[line width=1pt, color=Set1-A] table[x=epoch, y=train_loss_total, col sep=comma]{./figs/data/SGD0.01_beta5_NN1.csv};
        
        \addplot[line width=1pt, color=Set1-B] table[x=epoch, y=bc_loss_train, col sep=comma]{./figs/data/SGD0.01_beta5_NN1.csv};
        \addplot[line width=1pt, color=Set1-C] table[x=epoch, y=bulk_loss_train, col sep=comma]{./figs/data/SGD0.01_beta5_NN1.csv};
        \addplot[line width=1pt, color=Set1-D] table[x=epoch, y=bcp_loss_train, col sep=comma]{./figs/data/SGD0.01_beta5_NN1.csv};

        \addplot[line width=1pt, color=Set1-A, densely dotted] table[x=epoch, y=test_loss_total, col sep=comma]{./figs/data/SGD0.01_beta5_NN1.csv};
        
        \addplot[line width=1pt, color=Set1-B, densely dotted] table[x=epoch, y=bc_loss_test, col sep=comma]{./figs/data/SGD0.01_beta5_NN1.csv};
        \addplot[line width=1pt, color=Set1-C, densely dotted] table[x=epoch, y=bulk_loss_test, col sep=comma]{./figs/data/SGD0.01_beta5_NN1.csv};
        \addplot[line width=1pt, color=Set1-D, densely dotted] table[x=epoch, y=bcp_loss_test, col sep=comma]{./figs/data/SGD0.01_beta5_NN1.csv};

    \end{axis}
\end{tikzpicture}}}
    \\
    \subfloat[$\beta=15$, Network 1]{\adjustbox{width=0.42\linewidth,valign=b}{\begin{tikzpicture}
    \begin{axis}
    [
        axis line style={latex-latex},
        axis y line=left,
        axis x line=left,
        width=9cm,
        height=6cm,
        xmode=linear,
        ymode=log,
        xlabel = {Epoch},
        ylabel = {Loss},
        xmin = 0, xmax = 5200,
        xtick = {0,1000,2000,3000,4000,5000},
        ymin = 1e-5, ymax = 1,
        ytick = {1e0,1e-1,1e-2,1e-3,1e-4,1e-5,1e-6,1e-7},
        legend cell align={left},
        legend style={font=\scriptsize, at={(0.97, 0.97)},anchor=north east},
        %axis line style={draw=none},
        %tick style={draw=none},
        %x tick label style={/pgf/number format/.cd, fixed, fixed zerofill, precision=0, /tikz/.cd},
        %y tick label style={/pgf/number format/.cd, fixed, fixed zerofill, precision=1, /tikz/.cd},
    ]
        
        \addplot[line width=1pt, color=Set1-A] table[x=epoch, y=train_loss_total, col sep=comma]{./figs/data/SGD0.0001_beta15_NN0.csv};
        
        \addplot[line width=1pt, color=Set1-B] table[x=epoch, y=bc_loss_train, col sep=comma]{./figs/data/SGD0.0001_beta15_NN0.csv};
        \addplot[line width=1pt, color=Set1-C] table[x=epoch, y=bulk_loss_train, col sep=comma]{./figs/data/SGD0.0001_beta15_NN0.csv};
        \addplot[line width=1pt, color=Set1-D] table[x=epoch, y=bcp_loss_train, col sep=comma]{./figs/data/SGD0.0001_beta15_NN0.csv};

        \addplot[line width=1pt, color=Set1-A, densely dotted] table[x=epoch, y=test_loss_total, col sep=comma]{./figs/data/SGD0.0001_beta15_NN0.csv};
        
        \addplot[line width=1pt, color=Set1-B, densely dotted] table[x=epoch, y=bc_loss_test, col sep=comma]{./figs/data/SGD0.0001_beta15_NN0.csv};
        \addplot[line width=1pt, color=Set1-C, densely dotted] table[x=epoch, y=bulk_loss_test, col sep=comma]{./figs/data/SGD0.0001_beta15_NN0.csv};
        \addplot[line width=1pt, color=Set1-D, densely dotted] table[x=epoch, y=bcp_loss_test, col sep=comma]{./figs/data/SGD0.0001_beta15_NN0.csv};

    \end{axis}
\end{tikzpicture}}}
    \subfloat[$\beta=15$, Network 2]{\adjustbox{width=0.42\linewidth,valign=b}{\begin{tikzpicture}
    \begin{axis}
    [
        axis line style={latex-latex},
        axis y line=left,
        axis x line=left,
        width=9cm,
        height=6cm,
        xmode=linear,
        ymode=log,
        xlabel = {Epoch},
        ylabel = {Loss},
        xmin = 0, xmax = 5200,
        xtick = {0,1000,2000,3000,4000,5000},
        ymin = 1e-5, ymax = 1,
        ytick = {1e0,1e-1,1e-2,1e-3,1e-4,1e-5,1e-6,1e-7},
        legend cell align={left},
        legend style={font=\scriptsize, at={(0.97, 0.97)},anchor=north east},
        %axis line style={draw=none},
        %tick style={draw=none},
        %x tick label style={/pgf/number format/.cd, fixed, fixed zerofill, precision=0, /tikz/.cd},
        %y tick label style={/pgf/number format/.cd, fixed, fixed zerofill, precision=1, /tikz/.cd},
    ]
        
        \addplot[line width=1pt, color=Set1-A] table[x=epoch, y=train_loss_total, col sep=comma]{./figs/data/SGD0.01_beta15_NN1.csv};
        
        \addplot[line width=1pt, color=Set1-B] table[x=epoch, y=bc_loss_train, col sep=comma]{./figs/data/SGD0.01_beta15_NN1.csv};
        \addplot[line width=1pt, color=Set1-C] table[x=epoch, y=bulk_loss_train, col sep=comma]{./figs/data/SGD0.01_beta15_NN1.csv};
        \addplot[line width=1pt, color=Set1-D] table[x=epoch, y=bcp_loss_train, col sep=comma]{./figs/data/SGD0.01_beta15_NN1.csv};

        \addplot[line width=1pt, color=Set1-A, densely dotted] table[x=epoch, y=test_loss_total, col sep=comma]{./figs/data/SGD0.01_beta15_NN1.csv};
        
        \addplot[line width=1pt, color=Set1-B, densely dotted] table[x=epoch, y=bc_loss_test, col sep=comma]{./figs/data/SGD0.01_beta15_NN1.csv};
        \addplot[line width=1pt, color=Set1-C, densely dotted] table[x=epoch, y=bulk_loss_test, col sep=comma]{./figs/data/SGD0.01_beta15_NN1.csv};
        \addplot[line width=1pt, color=Set1-D, densely dotted] table[x=epoch, y=bcp_loss_test, col sep=comma]{./figs/data/SGD0.01_beta15_NN1.csv};

    \end{axis}
\end{tikzpicture}}}
    \\
    \subfloat[$\beta=30$, Network 1]{\adjustbox{width=0.42\linewidth,valign=b}{\begin{tikzpicture}
    \begin{axis}
    [
        axis line style={latex-latex},
        axis y line=left,
        axis x line=left,
        width=9cm,
        height=6cm,
        xmode=linear,
        ymode=log,
        xlabel = {Epoch},
        ylabel = {Loss},
        xmin = 0, xmax = 5200,
        xtick = {0,1000,2000,3000,4000,5000},
        ymin = 1e-5, ymax = 1,
        ytick = {1e0,1e-1,1e-2,1e-3,1e-4,1e-5,1e-6,1e-7},
        legend cell align={left},
        legend style={font=\scriptsize, at={(0.97, 0.97)},anchor=north east},
        %axis line style={draw=none},
        %tick style={draw=none},
        %x tick label style={/pgf/number format/.cd, fixed, fixed zerofill, precision=0, /tikz/.cd},
        %y tick label style={/pgf/number format/.cd, fixed, fixed zerofill, precision=1, /tikz/.cd},
    ]
        
        \addplot[line width=1pt, color=Set1-A] table[x=epoch, y=train_loss_total, col sep=comma]{./figs/data/SGD0.001_beta30_NN0.csv};
        
        \addplot[line width=1pt, color=Set1-B] table[x=epoch, y=bc_loss_train, col sep=comma]{./figs/data/SGD0.001_beta30_NN0.csv};
        \addplot[line width=1pt, color=Set1-C] table[x=epoch, y=bulk_loss_train, col sep=comma]{./figs/data/SGD0.001_beta30_NN0.csv};
        \addplot[line width=1pt, color=Set1-D] table[x=epoch, y=bcp_loss_train, col sep=comma]{./figs/data/SGD0.001_beta30_NN0.csv};

        \addplot[line width=1pt, color=Set1-A, densely dotted] table[x=epoch, y=test_loss_total, col sep=comma]{./figs/data/SGD0.001_beta30_NN0.csv};
        \addplot[line width=1pt, color=Set1-B, densely dotted] table[x=epoch, y=bc_loss_test, col sep=comma]{./figs/data/SGD0.001_beta30_NN0.csv};
        \addplot[line width=1pt, color=Set1-C, densely dotted] table[x=epoch, y=bulk_loss_test, col sep=comma]{./figs/data/SGD0.001_beta30_NN0.csv};
        \addplot[line width=1pt, color=Set1-D,densely dotted] table[x=epoch, y=bcp_loss_test, col sep=comma]{./figs/data/SGD0.001_beta30_NN0.csv};

    \end{axis}
\end{tikzpicture}}}
    \subfloat[$\beta=30$, Network 2]{\adjustbox{width=0.42\linewidth,valign=b}{\begin{tikzpicture}
    \begin{axis}
    [
        axis line style={latex-latex},
        axis y line=left,
        axis x line=left,
        width=9cm,
        height=6cm,
        xmode=linear,
        ymode=log,
        xlabel = {Epoch},
        ylabel = {Loss},
        xmin = 0, xmax = 5200,
        xtick = {0,1000,2000,3000,4000,5000},
        ymin = 1e-5, ymax = 1,
        ytick = {1e0,1e-1,1e-2,1e-3,1e-4,1e-5,1e-6,1e-7},
        legend cell align={left},
        legend style={font=\scriptsize, at={(0.97, 0.97)},anchor=north east},
        %axis line style={draw=none},
        %tick style={draw=none},
        %x tick label style={/pgf/number format/.cd, fixed, fixed zerofill, precision=0, /tikz/.cd},
        %y tick label style={/pgf/number format/.cd, fixed, fixed zerofill, precision=1, /tikz/.cd},
    ]
        
        \addplot[line width=1pt, color=Set1-A] table[x=epoch, y=train_loss_total, col sep=comma]{./figs/data/SGD0.001_beta30_NN1.csv};
        
        \addplot[line width=1pt, color=Set1-B] table[x=epoch, y=bc_loss_train, col sep=comma]{./figs/data/SGD0.001_beta30_NN1.csv};
        \addplot[line width=1pt, color=Set1-C] table[x=epoch, y=bulk_loss_train, col sep=comma]{./figs/data/SGD0.001_beta30_NN1.csv};
        \addplot[line width=1pt, color=Set1-D] table[x=epoch, y=bcp_loss_train, col sep=comma]{./figs/data/SGD0.001_beta30_NN1.csv};

        \addplot[line width=1pt, color=Set1-A, densely dotted] table[x=epoch, y=test_loss_total, col sep=comma]{./figs/data/SGD0.001_beta30_NN1.csv};
        \addplot[line width=1pt, color=Set1-B, densely dotted] table[x=epoch, y=bc_loss_test, col sep=comma]{./figs/data/SGD0.001_beta30_NN1.csv};
        \addplot[line width=1pt, color=Set1-C, densely dotted] table[x=epoch, y=bulk_loss_test, col sep=comma]{./figs/data/SGD0.001_beta30_NN1.csv};
        \addplot[line width=1pt, color=Set1-D, densely dotted] table[x=epoch, y=bcp_loss_test, col sep=comma]{./figs/data/SGD0.001_beta30_NN1.csv};

    \end{axis}
\end{tikzpicture}}}
    \caption{\label{fig:gd_loss} Median loss versus epoch over 10 when using GD. \emph{Solid} line is the total, and the other lines show the separate contributions. \emph{dashed} for initial condition, \emph{dotted} for the bulk, and \emph{dash-dotted} for the periodic boundary.}
\end{figure}

Based on the performances in Fig.\,\ref{fig:error_beta} in the main text, we can divide the optimisers into two  groups:
\begin{itemize}
    \item \emph{LBFGS and ADAM are the optimisers that obtain the best results}. Although LBFGS converges faster, ADAM has the smallest error for the highest $\beta$ among all optimisers. Comparing train and test losses (\cref{fig:adam_loss,fig:lbfgd_loss}), we note that in both cases the generalisation errors (difference between test and train errors) become larger as $\beta$ increases.  Also, for large $\beta$, their losses are dominated by the periodic boundary contribution (last term  in Eq.\,\ref{eq:loss}), and test losses tend to overfit.
    \item \emph{The optimiser that performs worst is GD, followed by BBI}. In particular,  there is no learning for $\beta = 15$ and 30. In these cases,  the losses are dominated by the initial conditions (first term in Eq.\,\ref{eq:loss}), see Figs.\,\ref{fig:bbi_loss} and \ref{fig:gd_loss}.  The GD's poor performance is expected:  it has no momentum to be able to overcome saddle points and no mini-batches as a source of stochasticity that helps in escaping from local minima. On the other hand, explaining BBI's low performance requires further investigation as BBI's specialised hyperparameters offer a broad range of tuning possibilities \citep{de2022born}. In particular, training BBI for a longer time and exploring different strategies to trigger the bounces are possibilities worth exploring.
\end{itemize}

\clearpage

\subsection{Interpreting and visualizing training dynamics}
\label{app:training_dynamics}

This appendix aims to facilitate the understanding of the results shown in the main text, also providing a visualization when possible. Specifically, to better understand the behaviour of the trajectories, we calculate the Spearman correlation, $\rho$, between $\kappa$ and the MSE evaluated on the grid points. Spearman's correlation allows us to evaluate whether these quantities move together without fixing a functional form between the two. Furthermore, to understand the directions taken by the trajectory, we calculate the cosine similarity between two consecutive speeds:
\begin{equation}
    \cos (\theta_k) = \frac{\boldsymbol{\dot{\omega}}_k\cdot\boldsymbol{\dot{\omega}}_{k-1}}{\|\boldsymbol{\dot{\omega}}_k\| \|\boldsymbol{\dot{\omega}}_{k-1}\|} \qquad \mathrm{for} \qquad k=1,\dots,n_{\text{epochs}}\,.
\end{equation}

The graphs in \cref{figtab:all} show that the correlation between MSE and $\rho$ remains strongly negative (mainly $<-0.5$) in cases where the training reaches convergence (see, for example, the results where MSE $<0.01$). The third column in \cref{figtab:all} shows an example of the relation between $\kappa_\omega$ and MSE for $\beta = 1$ on the small (S) network. The colour of the dots varies from pale to intense following the number of evolution epochs. These examples show that the correlation between the curvature $\kappa_\omega$ and convergence remains negative even during the first stages of training. Therefore, the value of $\rho$ cannot be attributed only to some final spiralling in the last minimum of convergence. 
The rightmost column of \cref{figtab:all} shows an example (the same as the third column) of the relationship between the  ratio in magnitude and the cosine similarity of consecutive speed vectors. We see that while some optimisers show a preferred behaviour ($\theta_{\text{SGD}}$ tends to $\pi$ in the last stage of training, $\theta_{\text{Adam}}$ is nearly 0 before random dynamics kicks in, $\theta_{\text{BBI}}$ is always nearly 0), LBFGS does not show a preferred alignment with the gradient direction.

Finally, in \cref{fig:finalICMSE}, we show the relationship between the final values of curvature and MSE for each optimiser and $\beta$ value. In general, especially for gradient-based methods, it is possible to see an increase in the final intrinsic curvature value as the convergence increases (as MSE decreases).

\begin{table}[tbhp]
    \centering
    \caption{\label{figtab:all}Relation between convergence (MSE) and $\rho(\kappa_\omega, \text{MSE})$ or $\rho(\kappa_t, \text{MSE})$. Representative trajectory in $\kappa_\omega/\kappa_t$  and MSE space. Cosine similarity between consecutive speed vectors vs increase in magnitude in consecutive speeds.  The colour of the dots varies from pale to intense following the number of evolution epochs.}
    \begin{tabular}{M{1cm}M{3.1cm}M{3.1cm}M{3.1cm}M{3.1cm}N}
        \toprule
        & $\kappa_\omega$ & $\kappa_t$ & $\kappa$ vs. MSE & Trajectory \\\midrule
        GD & \includegraphics[width=3cm, height=3cm]{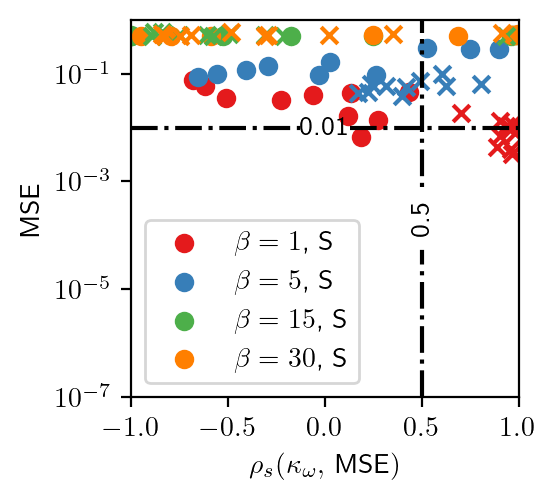} &    \includegraphics[width=3cm, height=3cm]{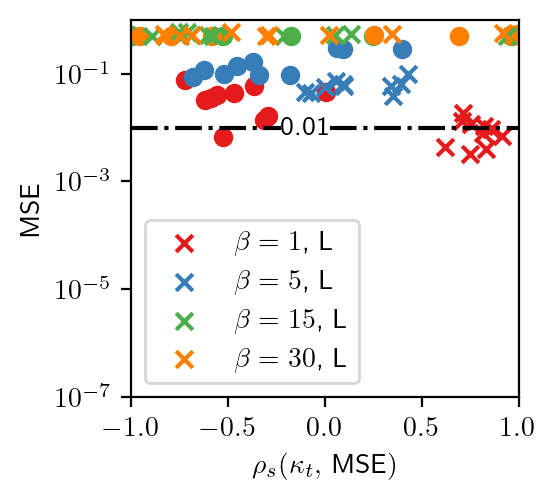} & \includegraphics[width=3cm, height=3cm]{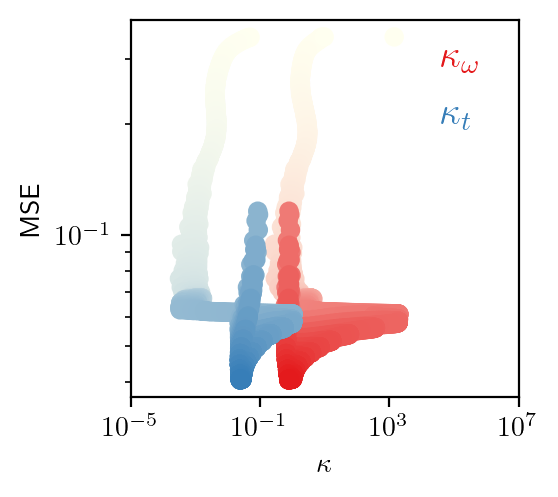} & \includegraphics[width=3cm, height=3cm]{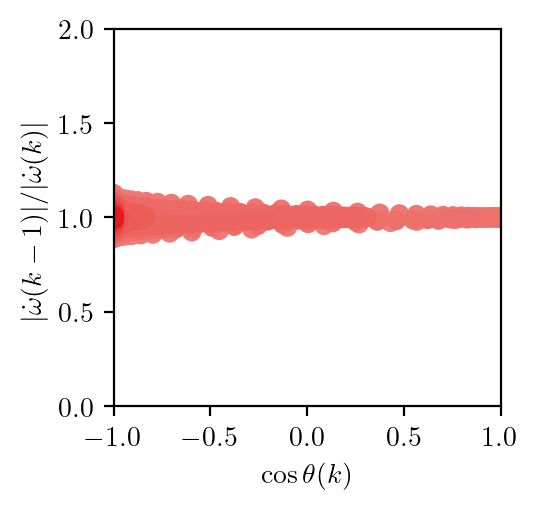} \\
        BBI & \includegraphics[width=3cm, height=3cm]{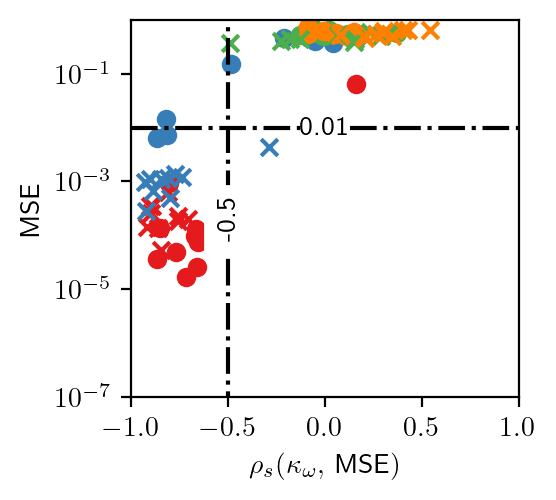} & \includegraphics[width=3cm, height=3cm]{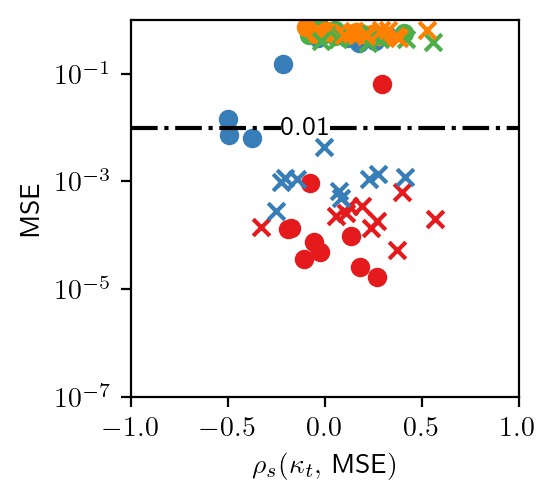} & \includegraphics[width=3cm, height=3cm]{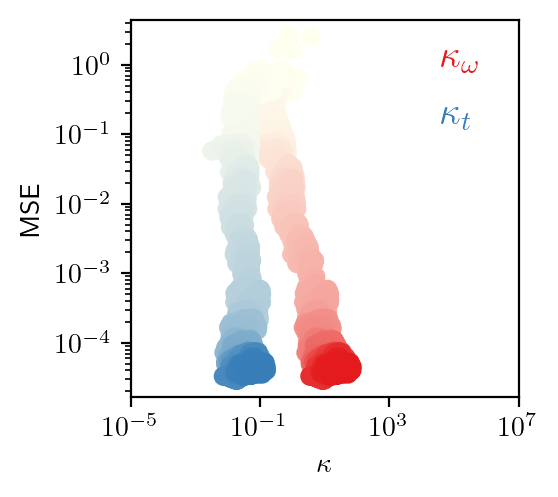} & \includegraphics[width=3cm, height=3cm]{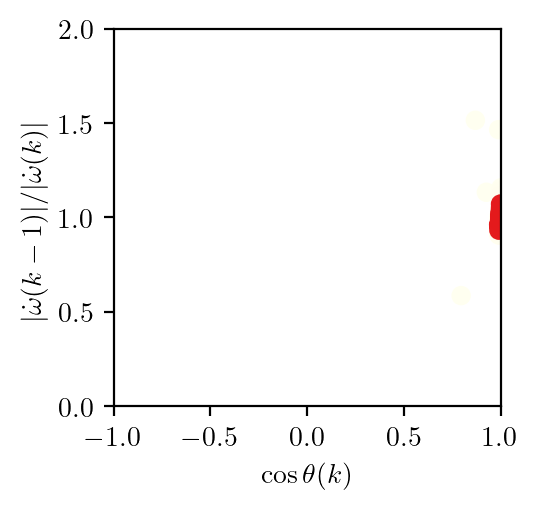}\\
        Adam & \includegraphics[width=3cm, height=3cm]{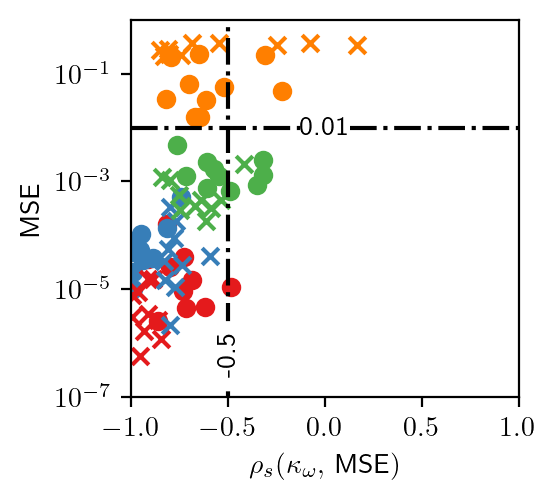} & \includegraphics[width=3cm, height=3cm]{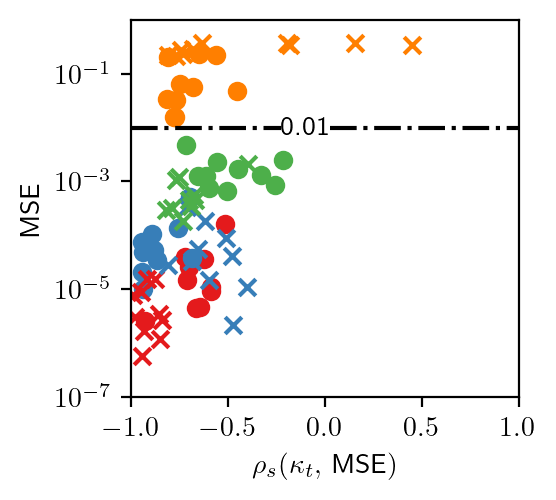} & \includegraphics[width=3cm, height=3cm]{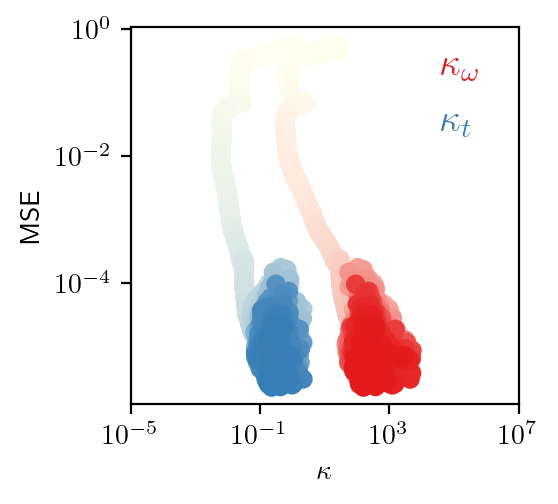} & \includegraphics[width=3cm, height=3cm]{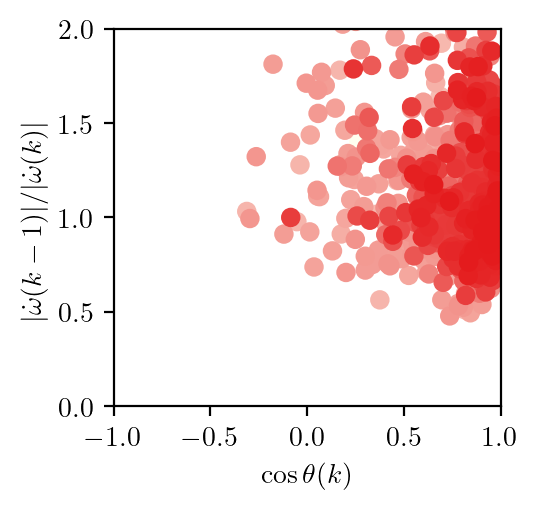} \\
        LBFGS & \includegraphics[width=3cm, height=3cm]{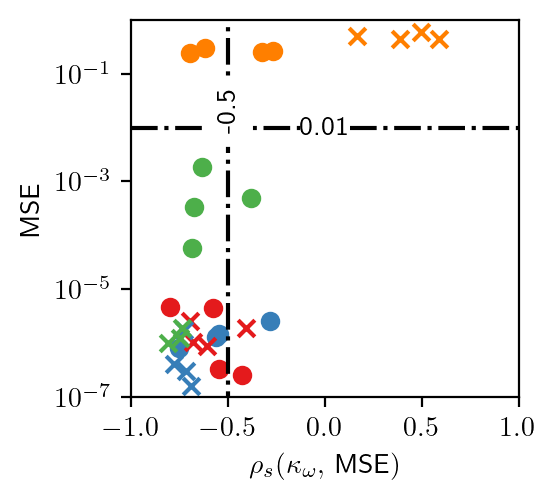} & \includegraphics[width=3cm, height=3cm]{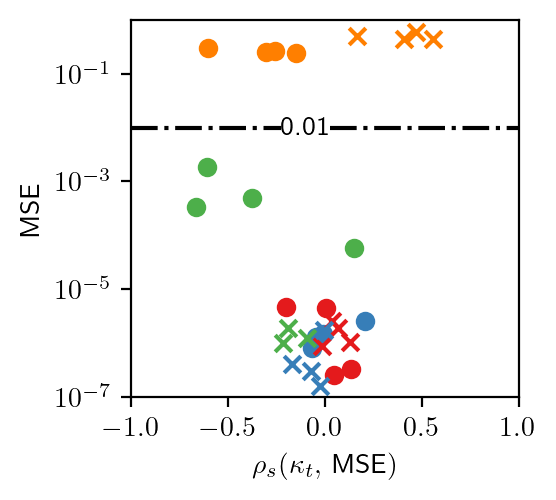} & \includegraphics[width=3cm, height=3cm]{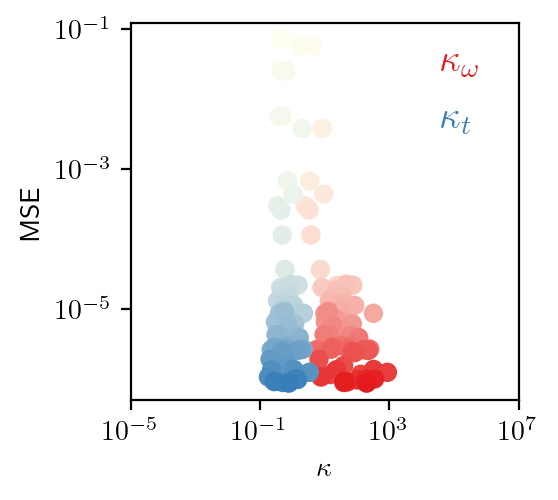} & \includegraphics[width=3cm, height=3cm]{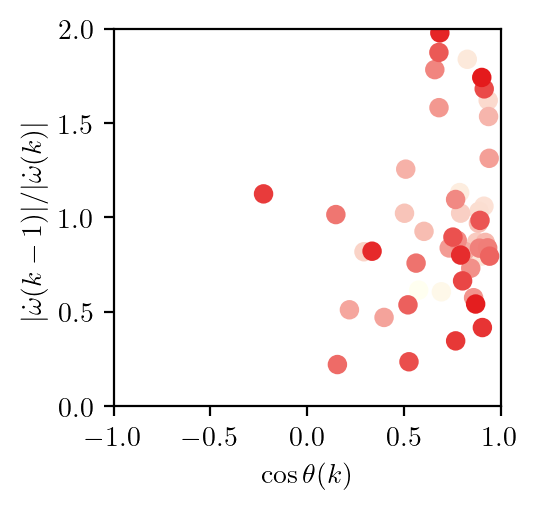} \\
    \bottomrule
  \end{tabular}
\end{table}

\begin{table}[tbhp]
    \centering
    \caption{\label{fig:finalICMSE}Effect of optimiser on the relation between final values of MSE and curvatures $\kappa_\omega$ and $\kappa_t$}
    \begin{tabular}{M{1cm}M{3.1cm}M{3.1cm}M{3.1cm}M{3.1cm}}
        \toprule
        & GD & BBI & Adam & LBFGS\\\midrule
        $\kappa_\omega$ & \includegraphics[width=3cm, height=3cm]{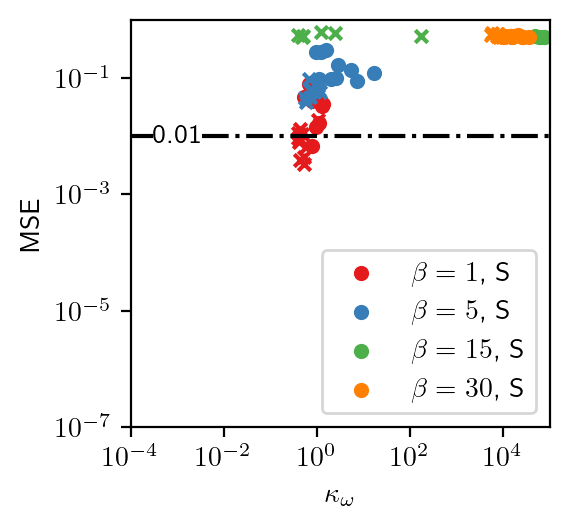} & \includegraphics[width=3cm, height=3cm]{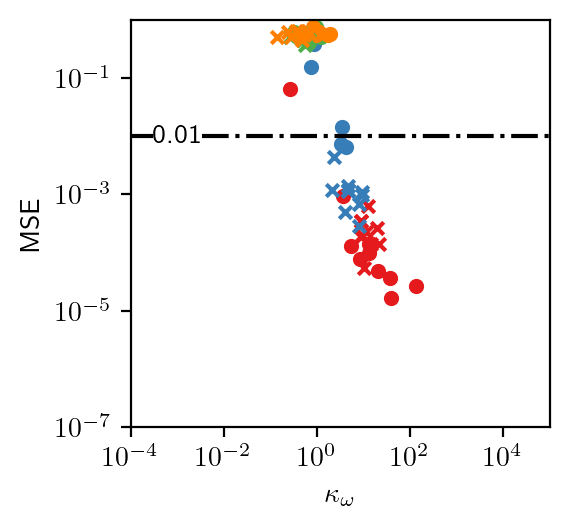} & \includegraphics[width=3cm, height=3cm]{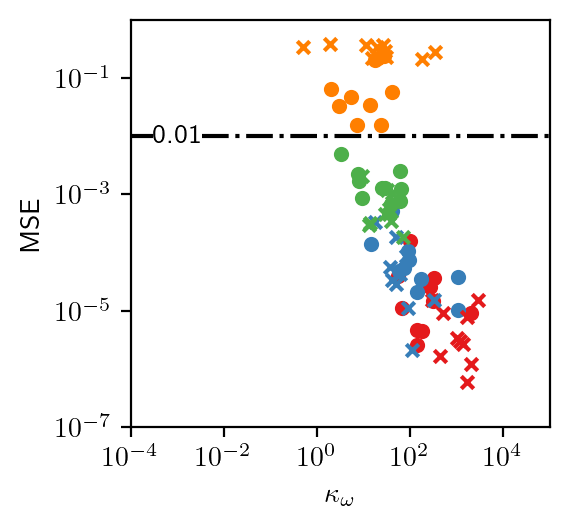} & \includegraphics[width=3cm, height=3cm]{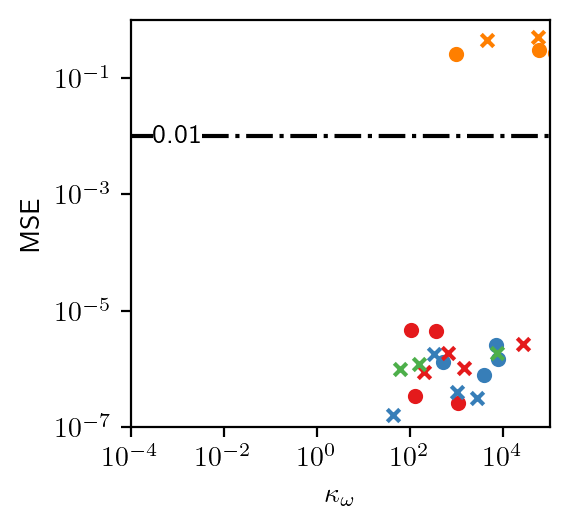} \\
        $\kappa_t$  &  \includegraphics[width=3cm, height=3cm]{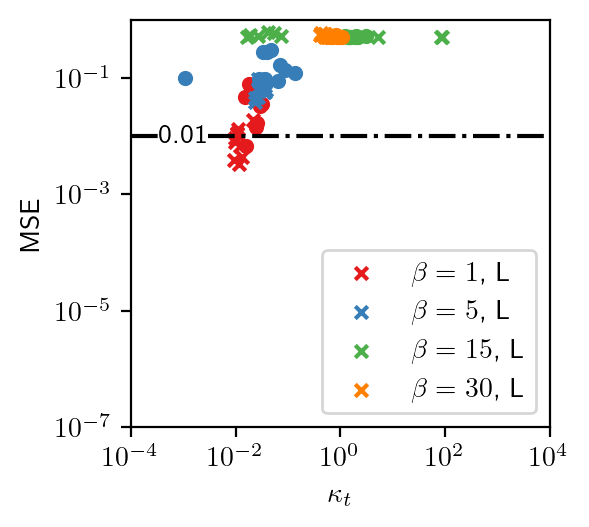} & \includegraphics[width=3cm, height=3cm]{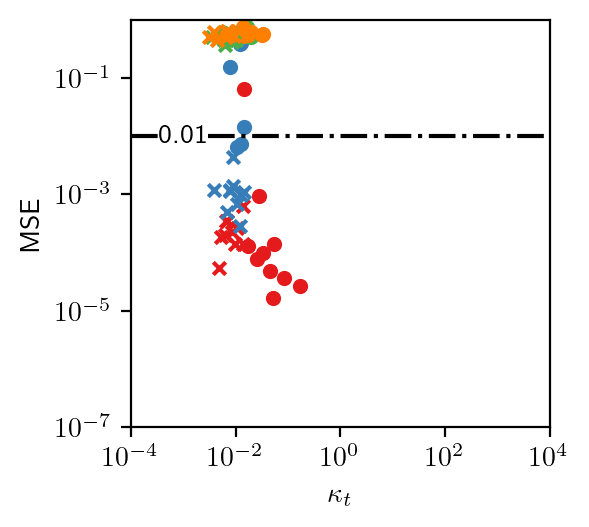} & \includegraphics[width=3cm, height=3cm]{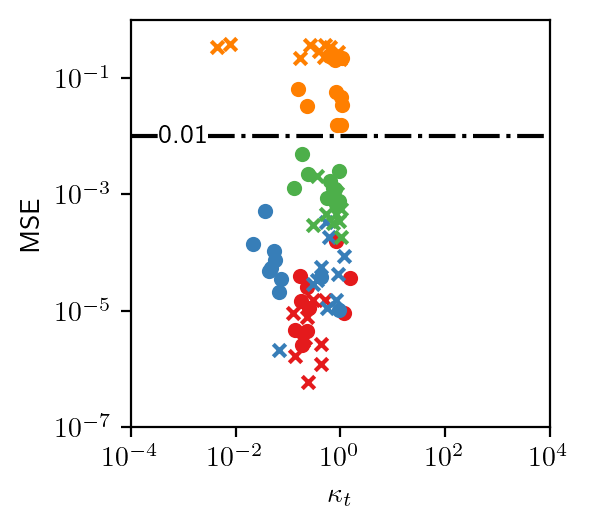} & \includegraphics[width=3cm, height=3cm]{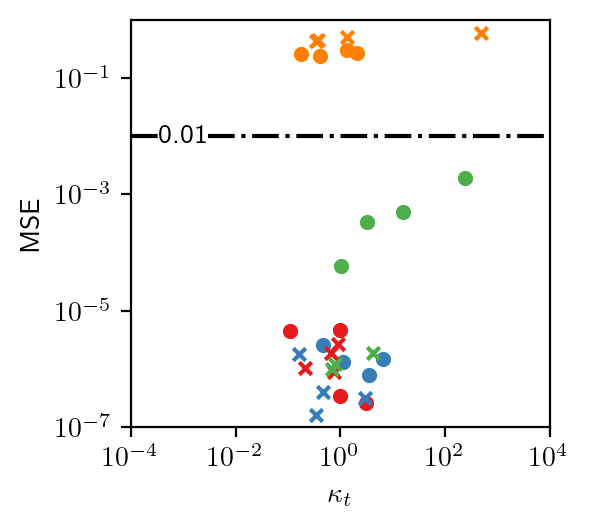}\\\bottomrule
    \end{tabular}
\end{table}

\end{document}